\documentclass[journal]{IEEEtran}
\usepackage{cite}
\pdfoutput=1
\ifCLASSINFOpdf
  \usepackage[pdftex]{graphicx}
\else
  \usepackage[dvips]{graphicx}
\fi
\usepackage[draft]{changes}
\setremarkmarkup{\textsuperscript{#2}}
\newcommand{\stkout}[1]{\ifmmode\text{\sout{\ensuremath{#1}}}\else\sout{#1}\fi}
\setdeletedmarkup{\stkout{#1}}
\usepackage{amsmath}
\usepackage{algorithmic}
\usepackage{algorithm}
\usepackage{array}
\usepackage{amsfonts}
\usepackage{amssymb}
\usepackage{bm}
\usepackage{color}
\usepackage{caption}
\usepackage[autostyle]{csquotes}
\ifCLASSOPTIONcompsoc
  \usepackage[caption=false,font=normalsize,labelfont=sf,textfont=sf]{subfig}
\else
  \usepackage[caption=false,font=footnotesize]{subfig}
\fi
\usepackage{url}

\begin{document}

\title{Structurally-sensitive Multi-scale Deep Neural Network for Low-Dose CT Denoising}

\author{Chenyu You, Qingsong Yang, Hongming Shan, Lars Gjesteby, Guang Li, Shenghong Ju, Zhuiyang Zhang, Zhen Zhao, Yi Zhang,~\IEEEmembership{Senior Member, IEEE}, Wenxiang Cong, and Ge Wang*,~\IEEEmembership{Fellow, IEEE}
\thanks{Asterisk indicates corresponding author.}
\thanks{C. You is with Departments of Bioengineering and Electrical Engineering, Stanford University, Stanford, CA, 94305 (e-mail: uniycy@stanford.edu)}
\thanks{Q. Yang, H. Shan, L. Gjesteby, G. Li, W. Cong, and G. Wang* are with Department of Biomedical Engineering, Rensselaer Polytechnic Institute, Troy, NY, 12180 (e-mail: yangq4@rpi.edu, shanh@rpi.edu, gjestl@rpi.edu, lig10@rpi.edu, congw@rpi.edu, wangg6@rpi.edu)}
\thanks{S. Ju, Z. Zhao are with Jiangsu Key Laboratory of Molecular and Functional Imaging, Department of Radiology, Zhongda Hospital, Medical School, Southeast University, Nanjing 210009, China (e-mail: jsh0836@hotmail.com, zhaozhen8810@126.com)}
\thanks{Y. Zhang is with the College of Computer Science, Sichuan University, Chengdu 610065, China (e-mail: yzhang@scu.edu.cn)}
\thanks{Z. Zhang is with Department of Radiology, Wuxi No.2 People's Hospital, Wuxi, 214000, China (e-mail: zhangzhuiyang@163.com)}
}

\maketitle

\begin{abstract}
Computed tomography (CT) is a popular medical imaging modality and enjoys wide clinical applications. At the same time, the x-ray radiation dose associated with CT scannings raises a public concern due to its potential risks to the patients. Over the past years, major efforts have been dedicated to the development of Low-Dose CT (LDCT) methods. However, the radiation dose reduction compromises the signal-to-noise ratio (SNR), leading to strong noise and artifacts that down-grade CT image quality. In this paper, we propose a novel 3D noise reduction method, called Structurally-sensitive Multi-scale Generative Adversarial Net (SMGAN), to improve the LDCT image quality. Specifically, we incorporate three-dimensional (3D) volumetric information to improve the image quality. Also, different loss functions for training denoising models are investigated. Experiments show that the proposed method can effectively preserve structural and textural information in reference to normal-dose CT (NDCT) images, and significantly suppress noise and artifacts. Qualitative visual assessments by three experienced radiologists demonstrate that the proposed method retrieves more information, and outperforms competing methods.
\end{abstract}

\begin{IEEEkeywords}
Machine Leaning, Low dose CT, Image denoising, Deep learning, Loss Function
\end{IEEEkeywords}

 \ifCLASSOPTIONpeerreview
 \begin{center} \bfseries EDICS Category: 3-BBND \end{center}
 \fi
%
\IEEEpeerreviewmaketitle

\section{Introduction}
\IEEEPARstart{X}{-ray} computed tomography (CT) is one of the most popular imaging modalities in clinical, industrial, and other applications~\cite{brenner2007computed}. Nevertheless, the potential risks (i.e., a chance to induce cancer and cause genetic damage) of ionizing radiation associated with medical CT scans cause a public concern~\cite{de2009projected}. Studies from the National Council on Radiation Protection and Measurements (NCRP) demonstrate a 600\% increase in medical radiation dose to the US population from 1980 to 2006, showing both great successes of the CT technology and an elevated alert to patients~\cite{schauer2009national}.

The main drawback of radiation dose reduction is to increase the image background noise, which could severely compromise diagnostic information. How to minimize the exposure to ionizing radiation while maintaining diagnostic utility of low-dose CT (LDCT) has been a challenge for researchers, who follows the well-known ALARA (as low as reasonably achievable) guideline~\cite{brenner2007computed}. Numerous methods were designed for LDCT noise reduction. These methods can be categorized as follows: (1)~\textit{Sinogram filtering-based techniques}~\cite{wang2005sinogram,wang2006penalized,balda2012ray,yang1996structure,liu2017discriminative,chen2008nonlocal}: these methods directly process projection data in the projection domain~\cite{balda2012ray}. The main advantage of these methods is computational efficiency. However, they may result in loss of structural information and spatial resolution~\cite{balda2012ray,manduca2009projection,yang1996structure}; 
(2)~\textit{Iterative reconstruction (IR)}~\cite{sidky2011constrained,de2004distance,whiting2006properties,elbakri2002statistical,tian2011low,liu2012adaptive,xu2012low,zhang2017tensor,sidky2008image,chen2013improving}: IR techniques may potentially produce high signal-to-noise ratio (SNR). However, these methods require a substantial computational cost and troublesome parametric turning; (3)~\textit{Image space denoising techniques}~\cite{ma2011low,li2014adaptive,chen2013improving,buades2005review,cheddad2012image,feruglio2010block,chen2014artifact,liu20163d}: these techniques can be performed directly on reconstructed images so that they can be applied across various CT scanners at a very low cost. Examples are non-local means-based filters~\cite{ma2011low,liu2012adaptive}, dictionary-learning-based K-singular value decomposition (K-SVD) method~\cite{chen2013improving} and the block-matching 3D (BM3D) algorithms~\cite{feruglio2010block,cheddad2012image}. Even though these algorithms greatly suppress noise and artifacts, edge blurring or resolution loss may persist in processed LDCT images.

Deep learning (DL) has recently received a tremendous attention in the field of medical imaging~\cite{wang2016perspective,wang2017machine}, such as brain image segmentation~\cite{zhang2015deep}, image registration~\cite{wang2017scalable,cao2018region}, image classification\cite{cattell2016classification}, and LDCT noise reduction~\cite{chen2017low,chen_zhang_kalra_lin_chen_liao_zhou_wang_2017,wolterink2017generative,yang2017low,kang2016deep,shan20183d,yang2017improving}. For example, Chen~\textit{et al.}~\cite{chen_zhang_kalra_lin_chen_liao_zhou_wang_2017} proposed a Residual Encoder-Decoder Convolutional Neural Network (REN-CNN) to predict NDCT images from noisy LDCT images. This method greatly reduces the background noise and artifacts. However, a limitation is that the results look blurry sometimes since the method targets minimizing the mean-squared error between the generated LDCT and corresponding NDCT images. To cope with this problem, the generative adversarial network (GAN)~\cite{goodfellow2014generative} offers an attractive solution. In the GAN, the generator $G$ learns to capture a real data distribution ${P_{r}}$ while the discriminator $D$ attempts to discriminate between the synthetic data distribution and the real counterpart. Note that the loss used in GAN, called the adversarial loss, measures the distance between the synthetic data distribution and the real one in order to improve the performance of $G$ and $D$ simultaneously. Originally, GAN uses the Jensen-Shannon (JS) divergence to evaluate the similarity of the two data distributions~\cite{goodfellow2014generative}. However, several problems exist in training GAN, such as unstable training and non-convergence. To address these issues, Arjovsky~\textit{et al.} introduced the Wasserstein distance instead of the Jensen-Shannon divergence to improve the neural network training~\cite{arjovsky2017wasserstein}. We will discuss more details on this aspect in Section~\ref{subsec:adversarial_loss}.

\begin{figure*}[!ht]
\begin{center}
\includegraphics[width=6.5in]{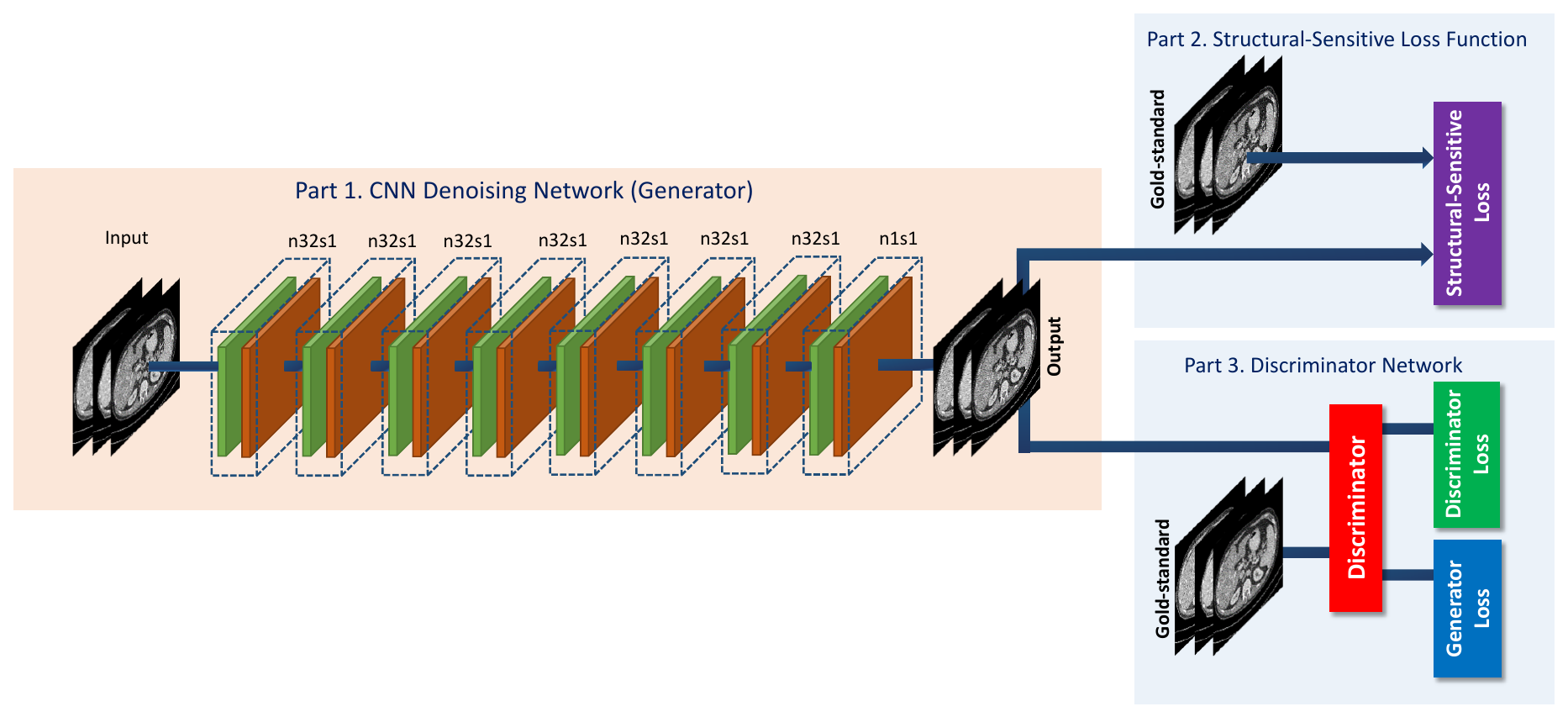}
\caption{The overall structure of the proposed SMGAN network. Note that the variable $n$ denotes the number of filters and $s$ denotes the stride size.}
\label{fig: overall}
\end{center}
\end{figure*}

In our previous work~\cite{yang2017low}, we first introduced the perceptual loss to capture perceptual differences between denoised  LDCT images and the reference NDCT images, providing the perceptually better results for clinical diagnosis at a cost of low scores in traditional image quality metrics. Since the traditional image quality metrics evaluate the generated images with reference to the gold-standard in generic ways, minimizing the perceptual loss does not ensure the results optimal in terms of the traditional image quality metrics. To address this discrepancy and inspired by the work in~\cite{wolterink2017generative,zhao2017loss}, here we propose a novel 3D clinical Structurally-sensitive Multi-scale Generative  Adversarial Network (SMGAN) to capture subtle structural features while maintaining high visual sensitivity. The proposed structurally-sensitive loss leverages a combination of adversarial loss\cite{arjovsky2017wasserstein}, perceptually-favorable structural loss, and pixel-wise $L_1$ loss. Moreover, to validate the diagnostic quality of images processed by our method, we report qualitative image assessments by three expert radiologists. Systematically, we demonstrate the feasibility and merits of mapping LDCT images to corresponding NDCT images in the GAN framework.

Our main contributions in this paper are summarized as follows:
\begin{enumerate}
  \item To keep the underlying structural information in LDCT images, we adopt a 3D CNN model as a generator based on WGAN which can enhance the image quality for better diagnosis.
  \item To measure the structural difference between generated LDCT images and the NDCT gold-standard, a structurally-sensitive loss is used to enhance the accuracy and robustness of the algorithm. Different from~\cite{yang2017low}, we replace the perceptual loss with a combination of $L_1$ loss and structural loss.
  \item To compare the performance of the 2D and the 3D models, we perform an extensive evaluation on their convergence rate and denoising performance.
\end{enumerate}

This paper is organized as follows: Section~\ref{sec:methods} introduces the proposed approach and analyzes the impact of each component loss function on the image quality. Section~\ref{sec:exp} presents the experimental design and results. Section~\ref{sec:discussion} discusses relevant issues. Finally, the concluding remarks and future plans are given in Section~\ref{table:notations}.

\section{Methods}
\label{sec:methods}

\subsection{Problem Inversion}
Assuming that $\bm{y} \in \mathbb{R}^{H \times W \times D}$ denotes the original LDCT image, and $\bm{x} \in \mathbb{R}^{H \times W \times D}$ denotes the corresponding NDCT image, the relationship between them can be expressed as:
\begin{equation}
\bm{y} = \bm{T}(\bm{x}) + \bm{\epsilon}
\end{equation}
where $T: \mathbb{R}^{H \times W \times D} \to \mathbb{R}^{H \times W \times D}$ is a generic noising process that degrades a real sample $\bm{x}$ of NDCT to a corresponding LDCT sample $\bm{y}$ in a non-linear way. $\bm{\epsilon}$ stands for the additive noise and unmodeled factors, and $H$, $W$, $D$ are height, width and depth respectively.

From another standpoint, considering that the real NDCT distribution $P_r$ is unknown, we focus on extracting information to recover desired images $\bm{x}$ from the noisy LDCT images $\bm{y}$. In general, the noise distribution in CT images is regarded as the mixture of Poisson quantum noise and Gaussian electronic noise\cite{fu2017comparison}. Compared with traditional denoising methods, the DL-based method is capable of effectively modeling any type of data distributions since the DL-based denoising model itself can be easily adapted to any practical noise model with statistical properties of typical noise distributions in a combination.
Therefore, the proposed DL-based denoising network is to solve the inverse problem $\bm{T^\dagger}\approx \bm{T}^{^{\,\_}1}$ to retrieve feasible images \bm{$\hat{x}$}, and the solution can be expressed as:
\begin{equation}
\bm{T^\dagger y} = \bm{\hat{x}}\approx\bm{x}
\end{equation}

As shown in Fig.\ref{fig: overall}, the overall network comprises three parts. Part 1 is the generator $G$, part 2 is the Structurally-Sensitive loss (SSL) function, and part 3 is the discriminator $D$. $G$ maps a volumetric LDCT image to the NDCT feature space, thereby estimating a NDCT image. The SSL function computes the structurally-sensitive dissimilarity which encodes multi-scale structural information. The loss computed by the SSL function aims to improve the ability of $G$ to generate realistic results. $D$ distinguishes a pair of synthetic and real NDCT images. If $D$ can identify the input image as \enquote{synthetic} or \enquote{real} correctly and tell us the discrepancy between the estimated CT image and the corresponding real NDCT image, we will know if $G$ yields a high-quality estimation or not. With the indication from $D$, $G$ can optimize its performance. Also, $D$ can upgrade its ability as well. Hence, $G$ and $D$ are in competition: $G$ attempts to generate a convincing estimate to an NDCT image while $D$ aims to distinguish the estimated image from real NDCT images. See Sections~\ref{subsec:network} and~\ref{subsec:loss_layer} for more details. For your convenience, the summary of notations that we use in this paper is in Table~\ref{table:notations}.

\subsection{3D Spatial Information}
The advantages of using 3D spatial information are evident. Hence, volumetric imaging and 3D visualization have become standards in diagnostic radiology~\cite{calhoun1999three}. There is a large amount of 3D NDCT and LDCT volumetric images available in practice. However, most of the networks are of 2D-based architecture. With a 3D network architecture, adjacent cross-section slices from a 3D CT image volume exhibit strong spatial correlation which we can utilize to preserve more information than with 2D models. 

As mentioned above, here we use a 3D ConvNet as the generator and introduce a 3D Structurally-Sensitive loss (SSL) function. Accordingly, we extract 3D image patches and use a 3D filter instead of a 2D filter. The generator in our network takes 3D volumetric LDCT patches as the input and process them with 3D non-linear transform operations. For convenience and comparison, 2D and 3D denoising networks are referred to as SMGAN-2D and SMGAN-3D respectively. The details of the network architecture are in the following Section~\ref{subsec:network}.

\subsection{Network Structure}
\label{subsec:network}
Inspired by the studies in~\cite{wolterink2017generative,yang2017low}, we introduce our proposed SMGAN-3D network structure. First, in Section~\ref{subsec:3d_generator} we present the 3D generator $G$ which captures local anatomical features. Then, in Section~\ref{subsec:STL_calculator} we define the 3D SSL function which guides the learning process . Finally, we outline the 2.5D discriminator $D$ in Section~\ref{subsec:discriminator}.

\subsubsection{\textbf{3D CNN Generator}}
\label{subsec:3d_generator}
The generator $G$ consists of eight 3D convolutional (Conv) layers. The first 7 layers each has 32 filters, and the last layer has only 1 filter. The odd-numbered convolutional layers apply $3\times3\times1$ filters, while the even-numbered convolutional layers use $3\times3\times3$ filters. The size of the extracted 3D patches is $80\times 80\times 11$ as the input to our whole network; see Fig.~\ref{fig: overall}. Note that the variable $n$ denotes the number of the filters and $s$ denotes the stride size, which is the step size of the filer when moving across an image so that $n32s1$ stands for 32 feature maps with a unit stride. Furthermore, a pooling layer after each Conv layer may lead to loss of subtle textural and structural information. Therefore, the pooling layer is not applied in this network. The Rectified Linear Unit (ReLU)~\cite{nair2010rectified} is our activation function after each Conv layer.

\subsubsection{\textbf{Structurally-Sensitive Loss (SSL) Function}}
\label{subsec:STL_calculator}
The proposed 3D SSL function measures the patch-wise discrepancy between a 3D output from the 3D ConvNet and the 3D NDCT image in the spatial domain. This measure is back-propagated\cite{simonyan2014very} through the neural network to update the parameters of the network; see Section~\ref{subsec:loss_layer} for more details.

\subsubsection{\textbf{Discriminator}}
\label{subsec:discriminator}
The discriminator $D$ consists of six convolutional layers with $64$, $64$, $128$, $128$, $256$, and $256$ filters and the kernel size of $3\times 3$. Two fully-connected (FC) layers produce $1024$ and $1$ feature maps respectively. Each layer is followed by a leaky ReLU defined as $\max(0, x) - \alpha \max(0, -x)$~\cite{nair2010rectified}, where $\alpha$ is a small constant. A stride of one pixel is applied for odd-numbered Conv layers and a stride of two pixels for even-numbered Conv layers. The input fed to $D$ is of the size ~$64\times 64\times 3$, which comes from the output of $G$. The reason why we use a 2D filter in $D$ is to reduce the computational complexity. Since the adversarial loss between each two adjacent slices in one volumetric patch contribute equally to the weighted average in one iteration, it can be easily computed. Following the suggestion in~\cite{arjovsky2017wasserstein}, we do not use the sigmoid cross entropy layer in $D$.

\subsection{Loss Functions for Noise Reduction}
\label{subsec:loss_layer}
In this sub-section, we evaluate the impact of different loss functions on LDCT noise reduction. This justifies the use of a hybrid loss function for optimal diagnostic quality.

\subsubsection{\bm{$L_2$}~\textbf{loss}}
\label{subsec:mse_loss}
The $L_2$ loss can efficiently suppress the background noise, but it could make the denoised results unnatural and blurry. This is expected due to its regression-to-mean nature~\cite{zhao2017loss,gulrajani2017improved}. Furthermore, the $L_2$ loss assumes that background noise is white Gaussian noise, which is independent of local image features~\cite{wang2004image} and not desirable for LDCT imaging. 

The formula of $L_{2}$ loss is expressed as:
\begin{equation}
L_{2} = \frac{1}{HWD}||G(\bm{y})-\bm{x}||_2^2
\label{eq: L2_loss}
\end{equation} 
where $H$, $W$, $D$ stand for the height, width, and depth of a 3D image patch respectively, $\bm{x}$ denotes the gold-standard (NDCT), and $G(\bm{y})$ represents the generated result from the source (LDCT) image $\bm{y}$. It is worth noting that since the $L_2$ loss has appealing properties of differentiability, convexity, and symmetry, the mean squared error (MSE) or $L_2$ loss is still a popular choice in denoising tasks\cite{wang2009mean}.

\subsubsection{\bm{$L_1$}~\textbf{Loss}}
\label{subsec:l1_loss}
The $L_1$ and $L_2$ losses are both the mean-based measures, the impacts of these two loss functions are different on denoising results. Compared with the $L_2$ loss, the $L_1$ loss does not over-penalize large differences or tolerate small errors between denoised and gold-standard images. Thus, the $L_1$ loss can alleviate some drawbacks of the $L_2$ loss we mentioned earlier. Additionally, the $L_1$ loss enjoys the same fine characteristics as $L_2$ loss except for the differentiability.

The formula for the $L_{1}$ loss is written as:
\begin{equation}
L_{1} = \frac{1}{HWD}|G(\bm{y})-\bm{x}|
\label{eq: L1_loss}
\end{equation} 
As shown in Figs.~\ref{fig: example1}\,-\ref{fig: example3_roi}, compared with the $L_2$ loss, the $L_1$ loss suppresses blurring, but does not help reduce blocky artifacts. For more details, see Section~\ref{sec:exp}.

\subsubsection{\textbf{Adversarial Loss}}
\label{subsec:adversarial_loss}
The Wasserstein distance with the regularization term was proposed in~\cite{gulrajani2017improved}, which is formulated as
\begin{equation}
L_{adv} = -\mathbb{E}[D(\bm{x})]+\mathbb{E}[D(\bm{z})] \\ +\lambda \mathbb{E}[(||\nabla_{\hat{\bm{x}}}D(\hat{\bm{x}})||_2-1)^2]
\end{equation}
where the first two terms are for the Wasserstein distance, and the third term implements the gradient penalty. Note that $\bm{z}$ denotes $G(\bm{y})$ for brevity. $\hat{\bm{x}}$ is uniformly sampled along the straight line between a pair of points sampled from $G$ and corresponding NDCT images.

\subsubsection{\textbf{Structural Loss}}
\label{subsec:structural_loss}
Medical images contain strong feature correlations. For example, their voxels have strong inter-dependencies. The structural similarity index (SSIM)~\cite{wang2004image} and the multi-scale structural similarity index (MS-SSIM)~\cite{wang2003multiscale} are perceptually motivated metrics, and perform better in visual pattern recognition than mean-based metrics~\cite{wang2004image}. To measure the structural and perceptual similarity between two images, the SSIM~\cite{wang2004image} is formulated as follows: 
\begin{align}
  SSIM(\bm{x},\bm{z}) &= \frac{2\mu_x\mu_z + C_1}{\mu_x^2 + \mu_z^2+C_1}\ast \frac{2 \sigma _{xz} + C_2}{\sigma_x^2 + \sigma_z^2+C_2}\label{eq:SSIM1} \\  
  &= l(\bm{x},\bm{z})\ast cs(\bm{x},\bm{z})\label{eq:SSIM2}
\end{align}
where $C_1$,$\,C_2$ are constants and $\,\mu_x$,$\,\mu_z$,$\,\sigma_x$,$\,\sigma_z$,$\,\sigma _{xz}$ denote means, standard deviations and cross-covariance of the image pair $(\bm{x},\bm{z})$ from $G$ and the corresponding NDCT image respectively. $l(\bm{x},\bm{z})$, $cs(\bm{x},\bm{z})$ are the first term and second factor we defined in Eqn.~\ref{eq:SSIM1}.

The multiscale SSIM provides more flexibility for multi-scale analysis~\cite{wang2003multiscale}. The formula for MS-SSIM~\cite{wang2003multiscale} is expressed as:
\begin{equation}
MS\_SSIM(\bm{x},\bm{z})=\prod_{j=1}^M SSIM(\bm{x}_{j},\bm{z}_{j})
\end{equation}
where $\bm{x}_{j}$,$\,\bm{z}_{j}$ are the local image content at the $j^{th}$ level, and $M$ is the number of scale levels. Clearly, SSIM is a special case of MS-SSIM.

The formula for the structural loss (SL) is generally expressed as:
\begin{equation}
L_{_{SL}} = 1-MS\_SSIM(\bm{x},\bm{z})
\label{eq: structure_loss}
\end{equation}
Note that the loss can be easily back-propagated to update weights in the network, since it can be differentiated~\cite{zhao2017loss}.  

\subsubsection{\textbf{Objective Function}}
\label{subsec:objective_function}

As mentioned in the recent studies~\cite{zhao2017loss,yang2017low}, minimizing the $L_2$ loss leads to over-smoothed appearance. The adversarial loss in GAN may yield sharp images, but it does not exactly match the corresponding real NDCT images~\cite{yang2017low}. The perceptual loss computed by a VGG network~\cite{simonyan2014very} evaluates the perceptual differences between the generated images and real NDCT images in a high-level feature space instead of the voxel space. Since the VGG network is trained on a large dataset of natural images, not CT images, it may result in distortions of processed CT images. To tackle these issues, we propose to utilize different loss terms together for high image quality.

As revealed in~\cite{zhao2017loss}, the  $L_1$ loss allows noise suppression and SNR improvement. However, it blurs anatomical structures to some extent. In contrast, the structural loss discourages blurring and keeps high contrast resolution. To have the merits of both loss functions, the structural sensitive loss (SSL) is expressed as:
\begin{equation}
L_{_{SSL}}=\tau\times L_{_{SL}}+(1-\tau)\times L_{1}
\label{eq: loss_mix}
\end{equation}
where $\tau$ is the weighting factor to balance between structure preservation in the first term (from Eq.~\ref{eq: structure_loss}) and noise suppression in the second term (from Eq.~\ref{eq: L1_loss}). 

Nevertheless, the above-mentioned two losses may still miss some diagnostic features. Hence, the adversarial loss is incorporated to keep textural and structural features as much as possible. In summary, the overall objective function of SMGAN is expressed as:
\begin{equation}
L_{obj} = L_{_{SSL}} + \beta\times L_{adv}
\label{eq: loss_final}
\end{equation}
where $\beta$ is the weight for the adversarial loss.
In the last step of the network, we compare the difference between the output volume and the target volume, and then the error can be back-propagated for optimization~\cite{lecun1998gradient}. 

\section{Experiments and results}
\label{sec:exp}
\subsection{Experimental Datasets and Setup}
To show the effectiveness of the proposed network for LDCT noise reduction, we used a real clinical dataset, published by Mayo Clinic for the~\textit{2016 NIH-AAPM-Mayo Clinic Low Dose CT Grand Challenge}~\cite{lowdosectgrandchallenge}. The Mayo dataset consists of 2,378 normal dose CT (NDCT) and low dose (quarter dose) CT (LDCT) images from 10 anonymous patients. The reconstruction interval and slice thickness in the dataset were $0.8mm$ and $1.0mm$ respectively.

For~\textbf{limited data}, the denoising performance of DL-based methods depends on the size of the training datasets, so large-scale valid training datasets can improve the denoising performance. However, it is worth noting that the training image library may not contain many valid images. To enhance the performance of the network, the strategies we utilized are as follows. First of all, in order to improve generalization performance of the network and avoid over-fitting, we adopted the \enquote{10-fold cross validation} strategy. The original dataset was partitioned into 10 equal size subsets. Then, a single subset was used in turn as the validation subset and the rest of data were utilized for training. Moreover, considering the limited number of CT images, we applied the overlapping patches strategy because it can not only consider patch-wise spatial interconnections, but also significantly increase the size of the training patch dataset~\cite{xie2012image,dong2016image}.

For~\textbf{data preprocessing}, the original LDCT and NDCT images are of $512\times 512$ pixels. Since directly processing the entire patient images is computationally inefficient and infeasible, our denoising model was applied to image patches. First, we applied the overlapped sliding window with a sliding size of $1\times 1\times 1$ to obtain image patches and then randomly extracted 100,100 pairs of training patches and 5,100 pairs for validation from remaining patient images of the same size $80\times 80\times 11$. Then, the \enquote{10-fold cross validation} strategy is used to ensure the accuracy of the proposed algorithm. Next, the CT Hounsfield Unit (HU) scale was normalized to [0, 1] before the images were fed to the network. 

For~\textbf{qualitative comparison}, in order to validate the performance of our proposed methods (SMGAN-2D and SMGAN-3D), we compare them with eight state-of-the-art denoising methods, including CNN-L2~($L_2$-net), CNN-L1~($L_1$-net), structural-loss net~(SL-net), multi-scale structural-loss net~(MSL-net), WGAN, BM3D~\cite{feruglio2010block}, RED-CNN~\cite{chen_zhang_kalra_lin_chen_liao_zhou_wang_2017}, and WGAN-VGG~\cite{yang2017low}. Among these existing denoising methods, BM3D is a classical image space denoising algorithm. WGAN-VGG represents a 2D perceptual-loss-based network, and RED-CNN refers to a 2D pixel-wise network. Note that the parameter settings in these methods~\cite{yang2017low,chen_zhang_kalra_lin_chen_liao_zhou_wang_2017,feruglio2010block} had been followed per the suggestions from the original papers.

For~\textbf{quantitative comparison}, to evaluate the effectiveness of the proposed methods, three metrics were chosen to perform image quality evaluation, including peak signal-to-noise ratio (PSNR), structural similarity index (SSIM)~\cite{wang2003multiscale}, and root-mean-square error (RMSE).

\begin{figure}[!t]
\subfloat[\label{fig: rmse_loss_convergence}]{\centerline{\includegraphics[width=3in]{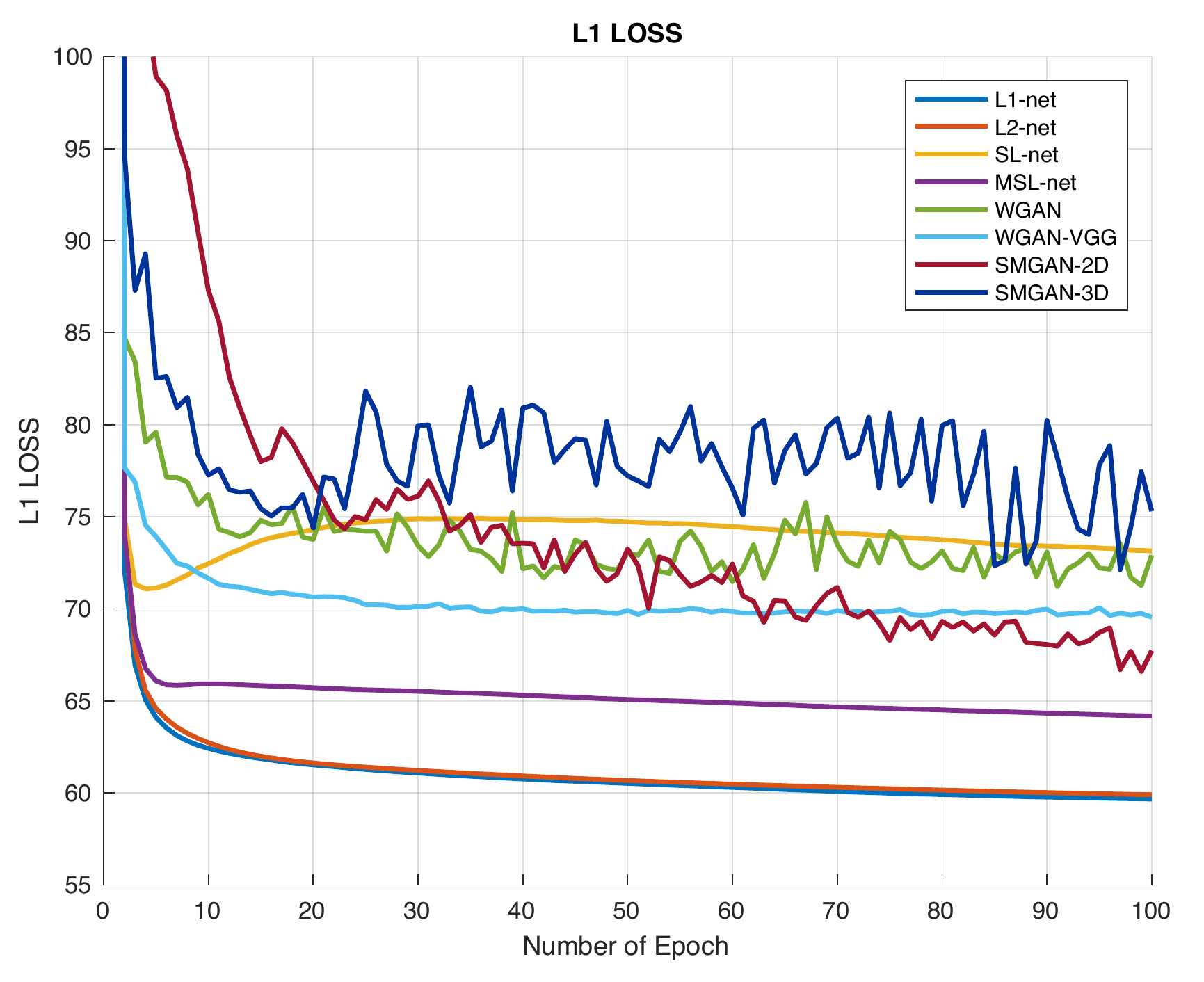}}}

\subfloat[\label{fig: structural_loss_convergence}]{\centerline{\includegraphics[width=3in]{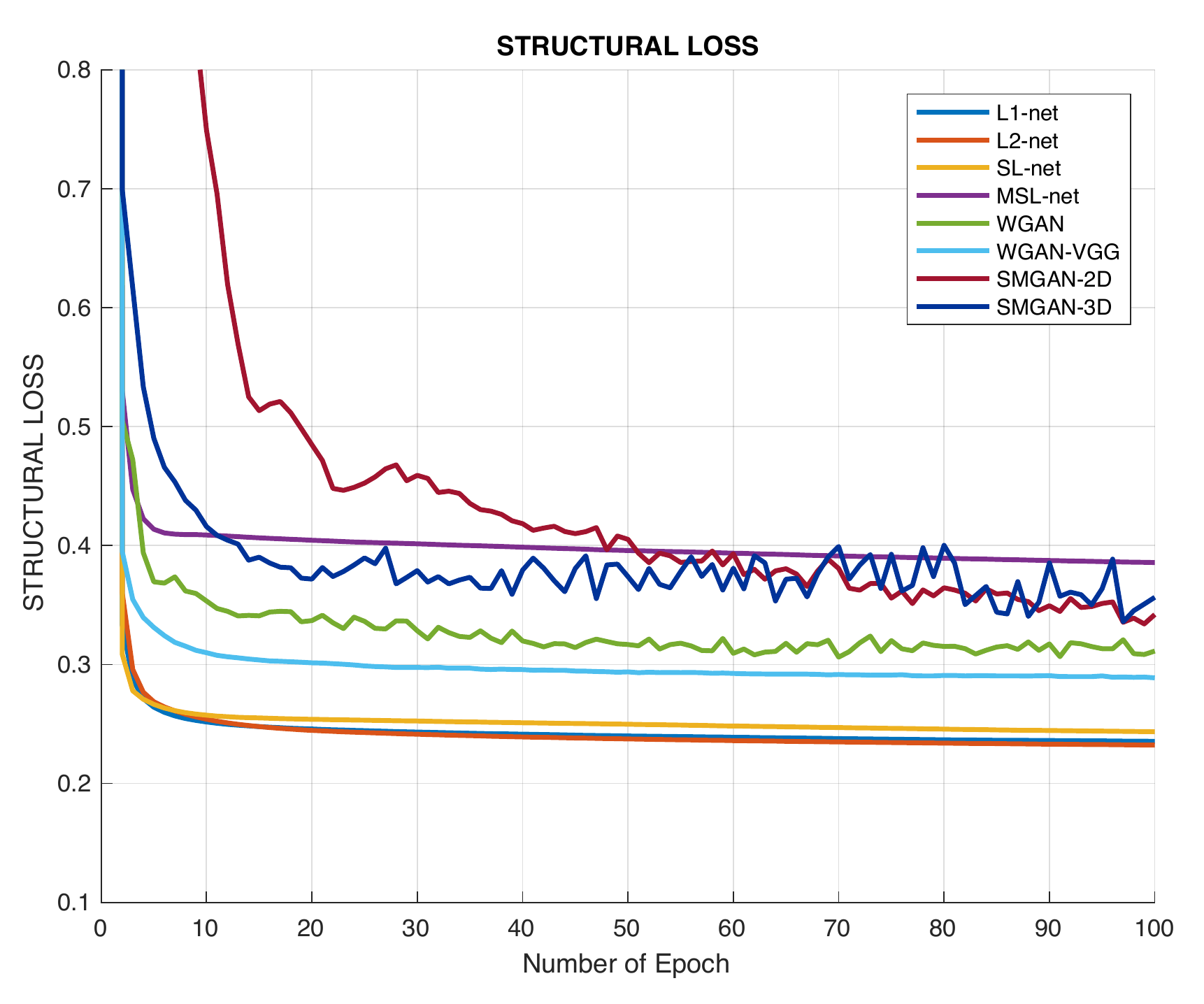}}}

\subfloat[\label{fig: wgan_loss_convergence}]{\centerline{\includegraphics[width=3in]{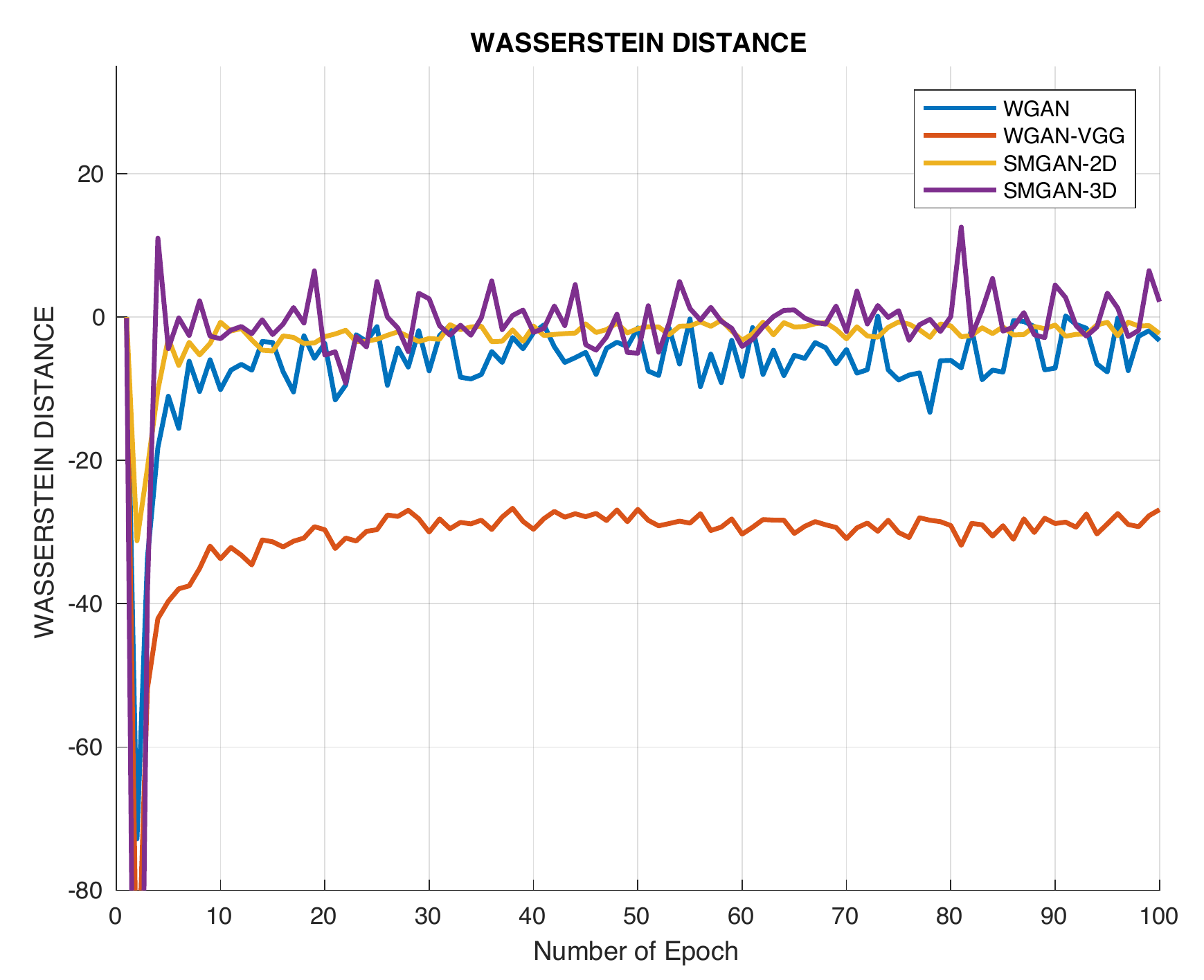}}}

\caption{Comparison of loss function value versus the number of epochs with respect to different algorithms. (a) L1 Loss, (b) Structural Loss, and (c) Wasserstein Distance curves.} 
\label{fig: convergence_data}
\end{figure}

\subsection{Parameter Selection}

In our experiments, the Adam optimization algorithm was implemented for our network training~\cite{kingma2014adam}. In the training phase, the mini-batch size was 64. The hyperparameter $\lambda$ for the balance between the Wasserstein distance and gradient penalty was set 10, per the suggestion from the original paper~\cite{arjovsky2017wasserstein}. The parameter $\beta$ for the trade-off between adversarial loss and mixture loss was set be $10^{-3}$. The parameter $\tau$ was set to 0.89. The slope of the leaky ReLu activation function was set to 0.2. The networks are implemented in the TensorFlow~\cite{abadi2016tensorflow} on an NVIDIA Titan Xp GPU.

\begin{figure*}[!t]
\centering
\subfloat[Full Dose FBP]{\includegraphics[width=1.7in]{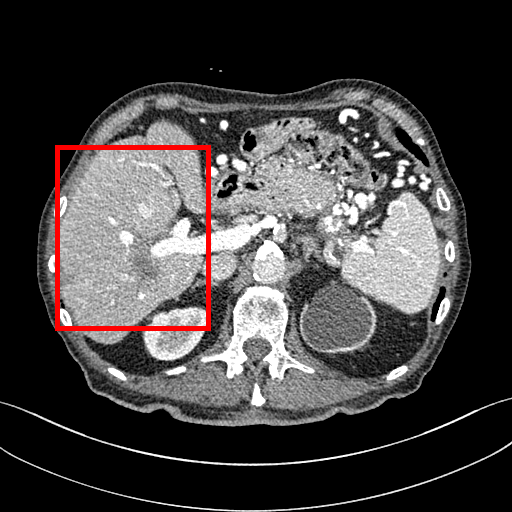}\label{fig: full1}}
\subfloat[Quarter Dose FBP]{\includegraphics[width=1.7in]{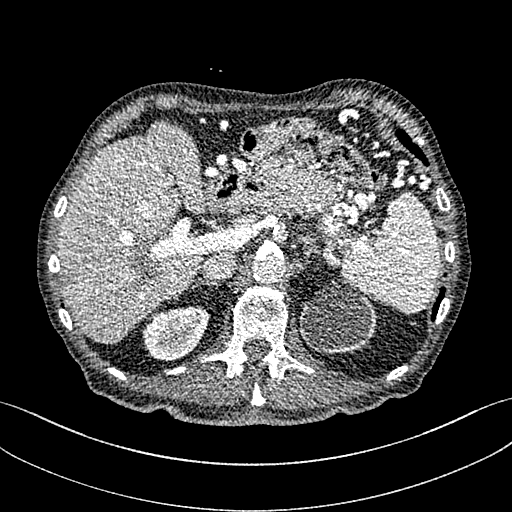}\label{fig: quarter1}}
\subfloat[CNN-L2]{\includegraphics[width=1.7in]{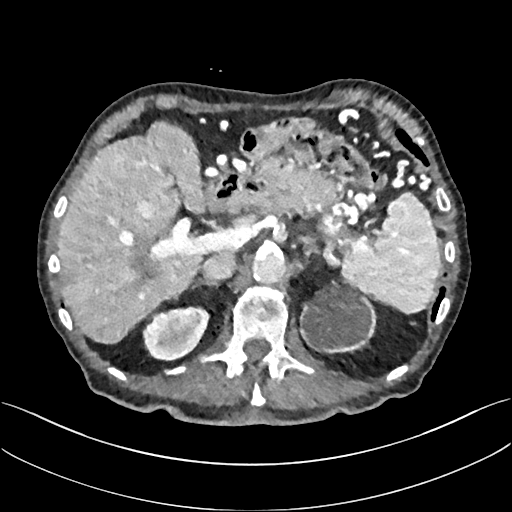}\label{fig: mse1}}
\subfloat[CNN-L1]{\includegraphics[width=1.7in]{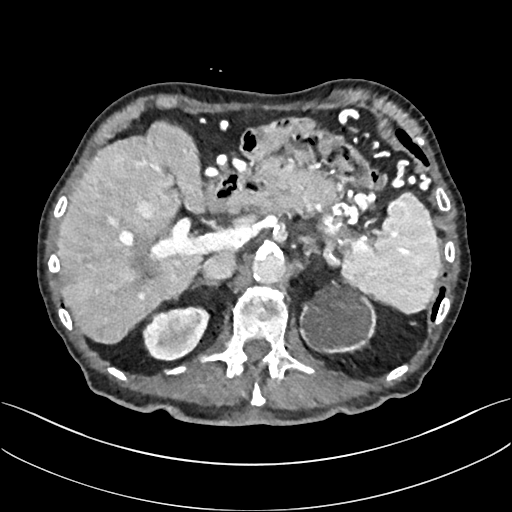}\label{fig: rmse1}}

\subfloat[CNN-SL]{\includegraphics[width=1.7in]{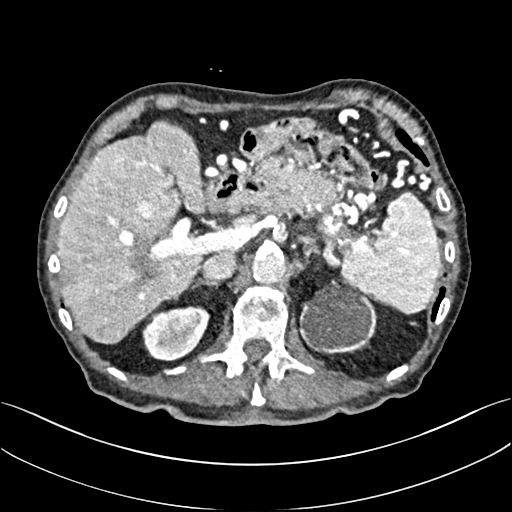}\label{fig: sl1}}
\subfloat[CNN-MSL]{\includegraphics[width=1.7in]{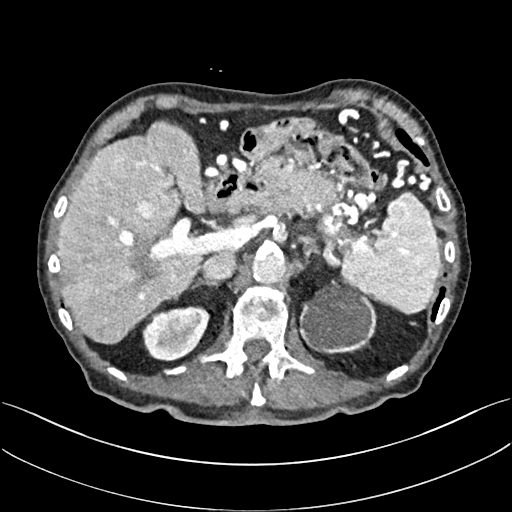}\label{fig: msl1}}
\subfloat[WGAN]{\includegraphics[width=1.7in]{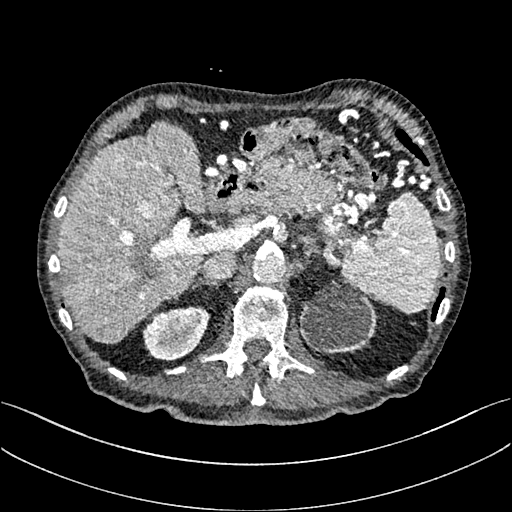}\label{fig: wgan1}}
\subfloat[BM3D]{\includegraphics[width=1.7in]{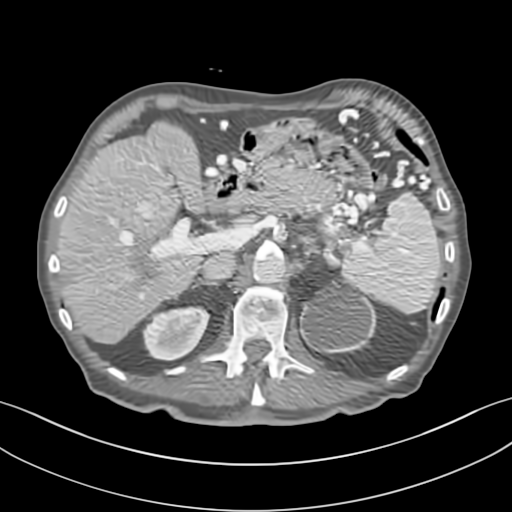}\label{fig: bm3d1}}

\subfloat[RED-CNN]{\includegraphics[width=1.7in]{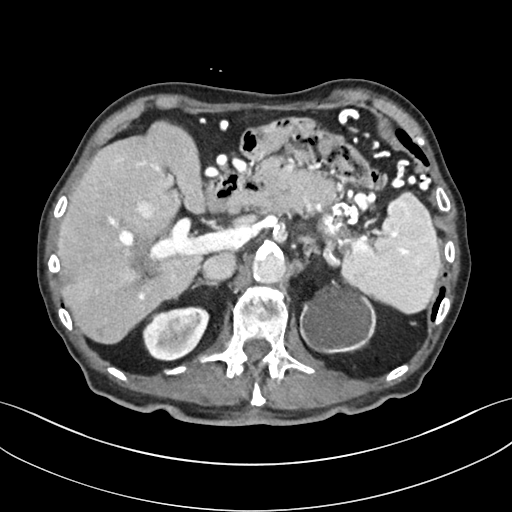}\label{fig: red_cnn1}}
\subfloat[WGAN-VGG]{\includegraphics[width=1.7in]{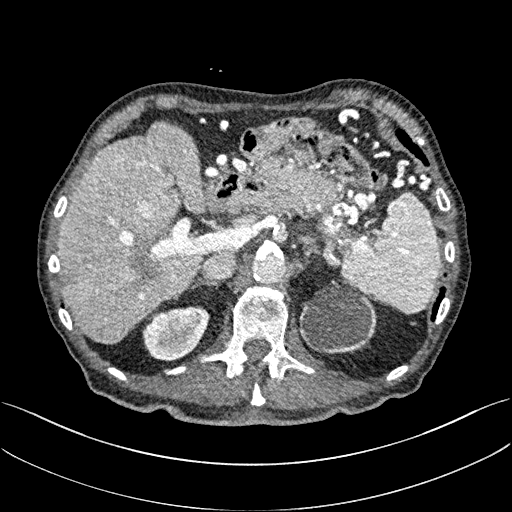}\label{fig: wgan_vgg1}}
\subfloat[SMGAN-2D]{\includegraphics[width=1.7in]{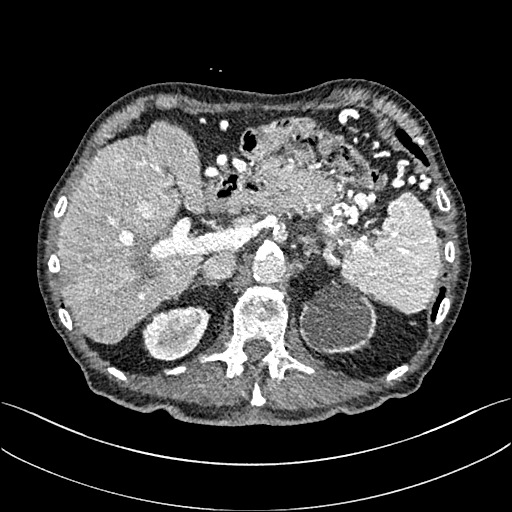}\label{fig: lsagan_2d1}}
\subfloat[SMGAN-3D]{\includegraphics[width=1.7in]{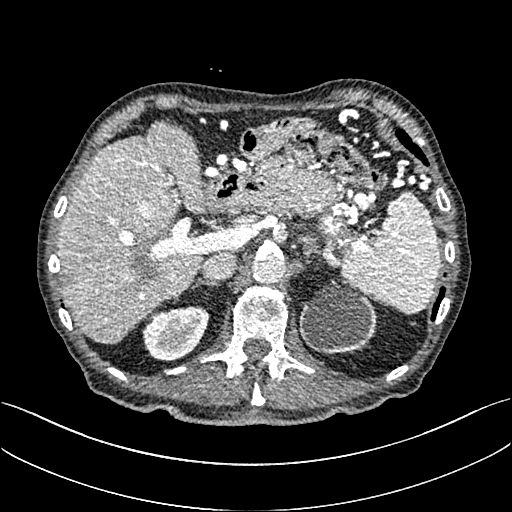}\label{fig: lsagan_3d1}}
\caption{Results from abdomen CT images.~(a) NDCT, (b) LDCT, (c) CNN-L2, (d) CNN-L1, (e) SL-net, (f) MSL-net, (g) WGAN (h) BM3D, (i) RED-CNN, (j) WGAN-VGG, (k) SMGAN-2D, and (l) SMGAN-3D. The red rectangle indicates the region zoomed in Fig.~\ref{fig: example1_roi}.
The display window is [-160, 240]HU.}
\label{fig: example1}
\end{figure*}

\begin{figure}[!t]
\centering
\subfloat[Full Dose FBP]{\includegraphics[width=0.80in]{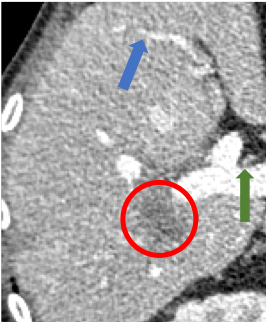}\label{fig: full1_roi}}\
\subfloat[Quarter Dose FBP]{\includegraphics[width=0.80in]{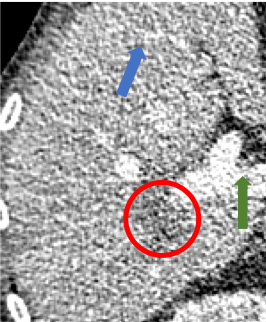}\label{fig: quarter1_roi}}\
\subfloat[CNN-L2]{\includegraphics[width=0.80in]{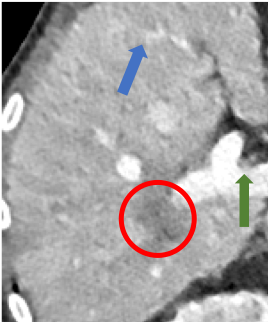}\label{fig: mse1_roi}}\
\subfloat[CNN-L1]{\includegraphics[width=0.80in]{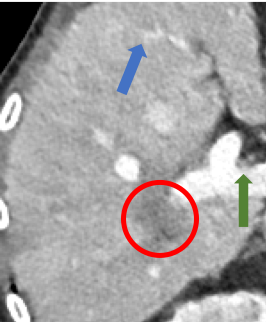}\label{fig: rmse1_roi}}

\subfloat[CNN-SL]{\includegraphics[width=0.80in]{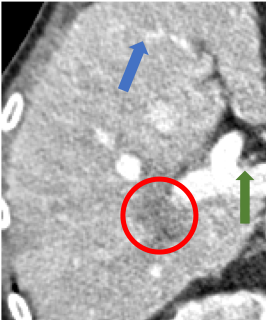}\label{fig: sl1_roi}}\
\subfloat[CNN-MSL]{\includegraphics[width=0.80in]{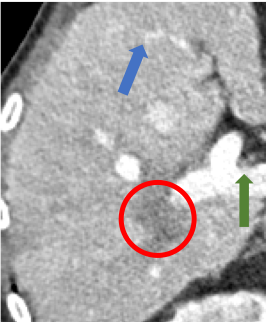}\label{fig: msl1_roi}}\
\subfloat[WGAN]{\includegraphics[width=0.80in]{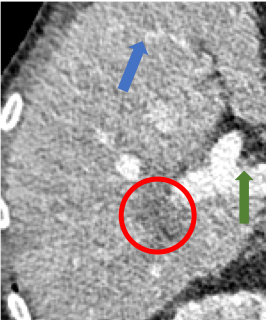}\label{fig: wgan1_roi}}\
\subfloat[BM3D]{\includegraphics[width=0.80in]{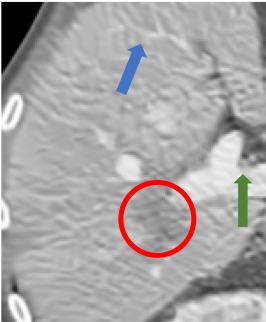}\label{fig: bm3d1_roi}}

\subfloat[RED-CNN]{\includegraphics[width=0.80in]{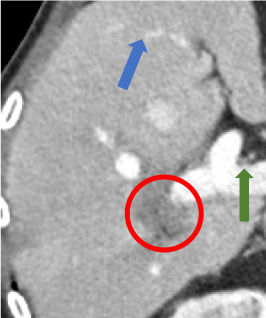}\label{fig: red_cnn1_roi}}\
\subfloat[WGAN-VGG]{\includegraphics[width=0.80in]{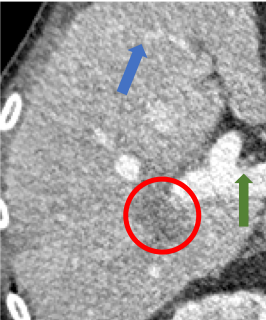}\label{fig: wgan_vgg1_roi}}\
\subfloat[SMGAN-2D]{\includegraphics[width=0.80in]{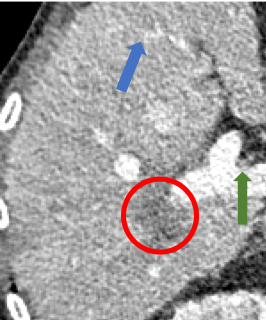}\label{fig: lsagan_2d1_roi}}\
\subfloat[SMGAN-3D]{\includegraphics[width=0.80in]{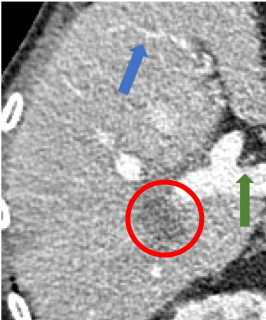}\label{fig: lsagan_3d1_roi}}
\caption{Zoomed parts of the region of interests~(ROIs) marked by the red rectangle in Fig.~\ref{fig: example1}. (a) NDCT, (b) LDCT, (c) CNN-L2, (d) CNN-L1, (e) SL-net, (f) MSL-net, (g) WGAN, (h) BM3D, (i) RED-CNN, (j) WGAN-VGG, (k) SMGAN-2D and (l) SMGAN-3D. The red circle indicates the metastasis and the green and blue arrows indicate two subtle structure parts. The display window is [-160,240]HU.}
\label{fig: example1_roi}
\end{figure}
\subsection{Network Convergence}

To examine the robustness of different denoising algorithms, ten methods corresponding to the $L_{1}$ loss $(L_{1})$, structural loss (SL), and Wasserstein distance were separately trained in the same settings as that for SMGAN-3D. Note that the parameter settings of RED-CNN, WGAN-VGG, and BM3D from the original papers had been followed~\cite{chen_zhang_kalra_lin_chen_liao_zhou_wang_2017,yang2017low,feruglio2010block}. In addition, the size of the input patches of the 2D network is $80\times 80$ while our proposed 3D model uses training patches with the size of $80\times 80\times 11$. We calculated the averaged loss value achieved by different methods versus the number of epochs as the measure of convergence in Fig.~\ref{fig: convergence_data}.

\begin{figure*}[!t]
\centering
\subfloat[Full Dose FBP]{\includegraphics[width=1.7in]{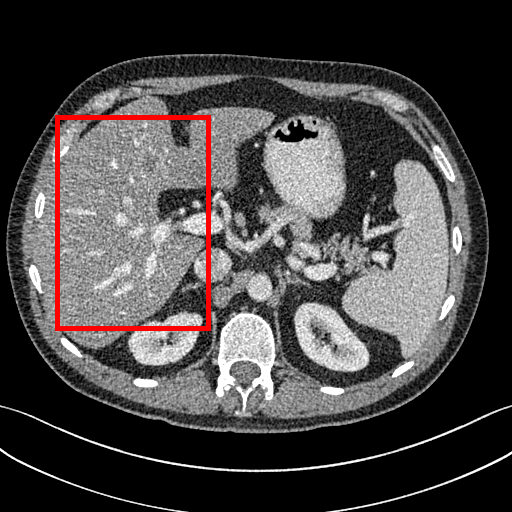}\label{fig: full2}}
\subfloat[Quarter Dose FBP]{\includegraphics[width=1.7in]{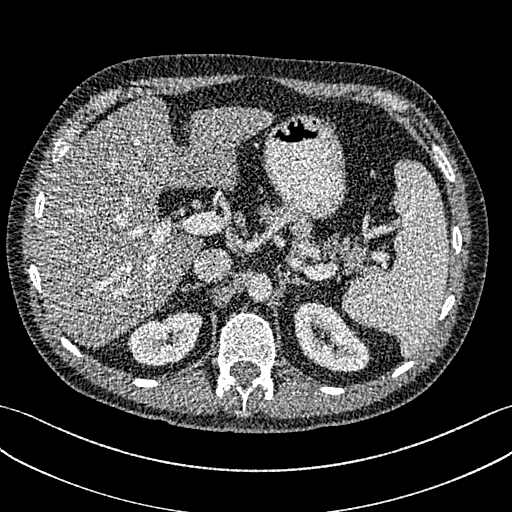}\label{fig: quarter2}}
\subfloat[CNN-L2]{\includegraphics[width=1.7in]{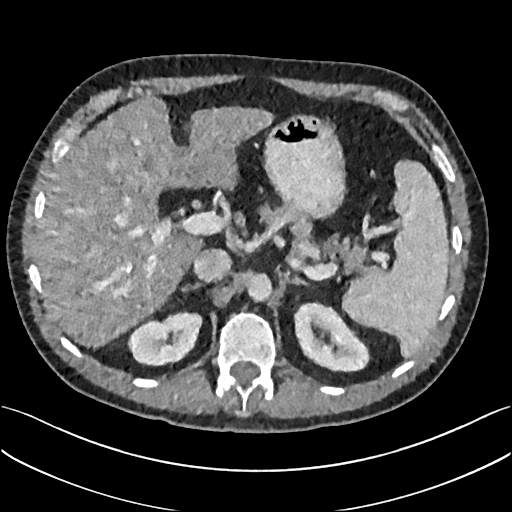}\label{fig: mse2}}
\subfloat[CNN-L1]{\includegraphics[width=1.7in]{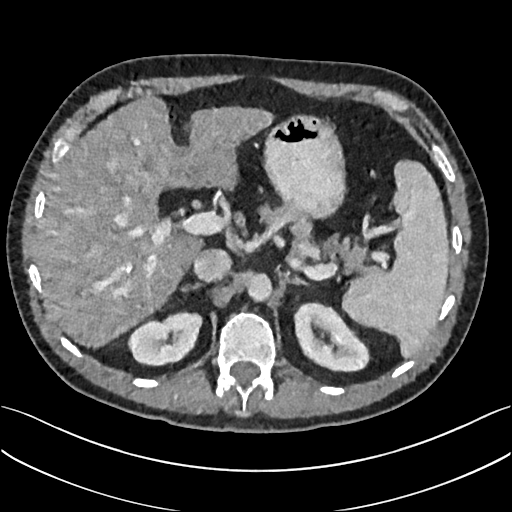}\label{fig: rmse2}}

\subfloat[CNN-SL]{\includegraphics[width=1.7in]{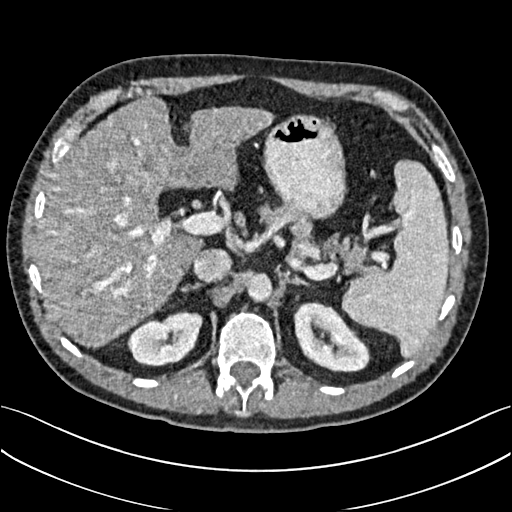}\label{fig: sl2}}
\subfloat[CNN-MSL]{\includegraphics[width=1.7in]{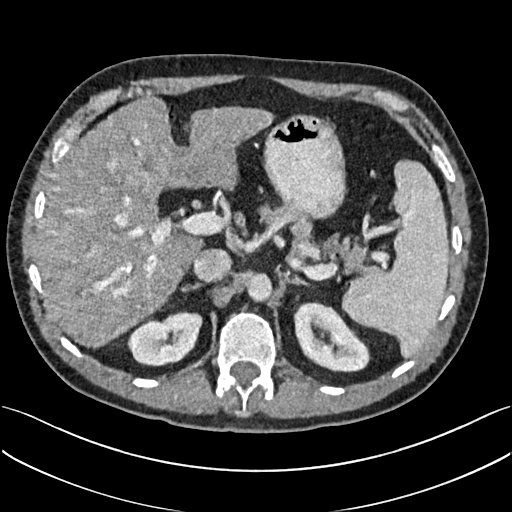}\label{fig: msl2}}
\subfloat[WGAN]{\includegraphics[width=1.7in]{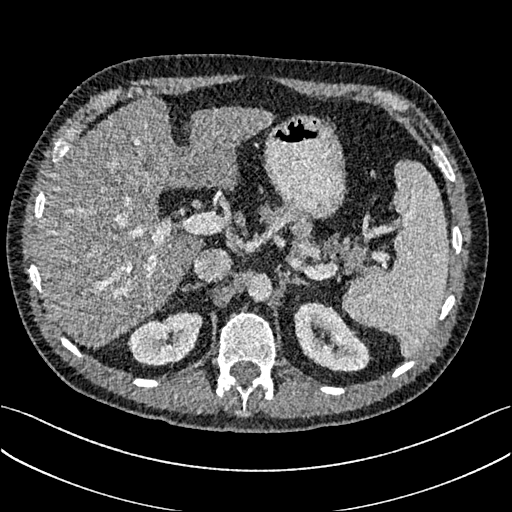}\label{fig: wgan2}}
\subfloat[BM3D]{\includegraphics[width=1.7in]{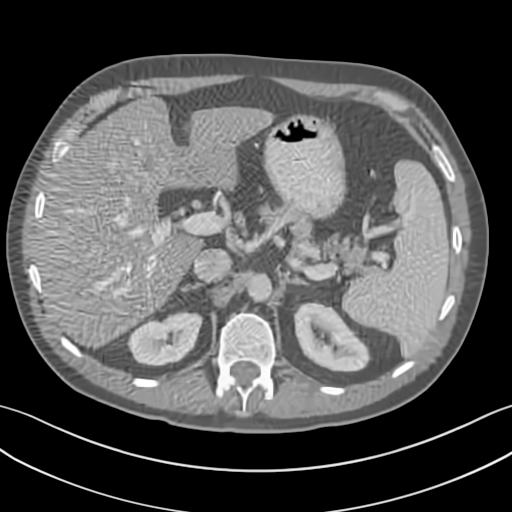}\label{fig: bm3d2}}

\subfloat[RED-CNN]{\includegraphics[width=1.7in]{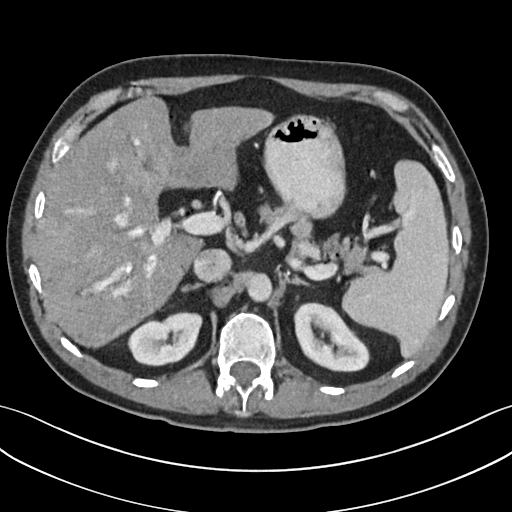}\label{fig: red_cnn2}}
\subfloat[WGAN-VGG]{\includegraphics[width=1.7in]{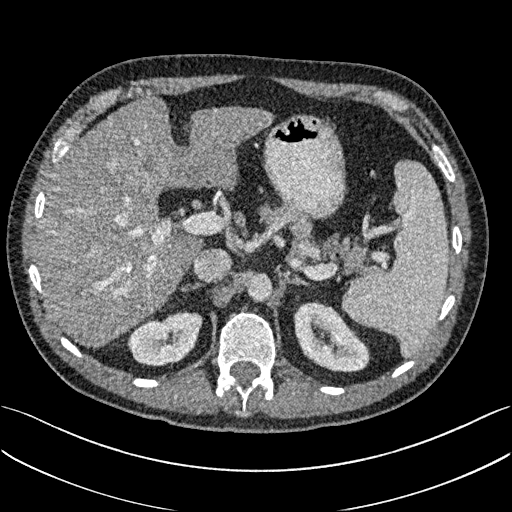}\label{fig: wgan_vgg2}}
\subfloat[SMGAN-2D]{\includegraphics[width=1.7in]{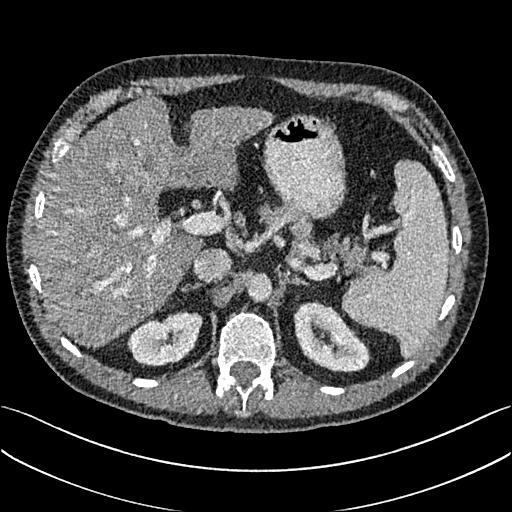}\label{fig: lsagan_2d2}}
\subfloat[SMGAN-3D]{\includegraphics[width=1.7in]{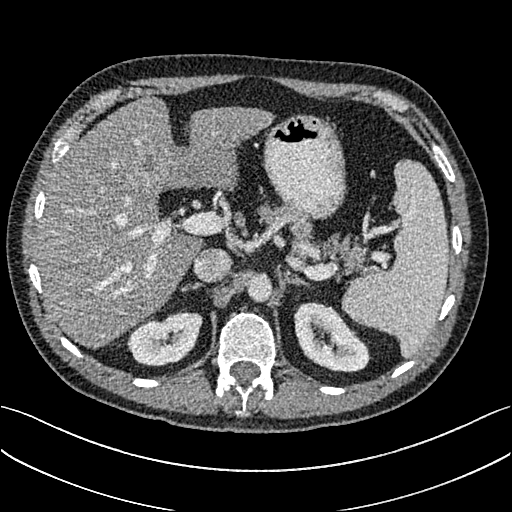}\label{fig: lsagan_3d2}}
\caption{Results from abdomen CT images.~(a) NDCT, (b) LDCT, (c) CNN-L2, (d) CNN-L1, (e) SL-net, (f) MSL-net, (g) WGAN (h) BM3D, (i) RED-CNN, (j) WGAN-VGG, (k) SMGAN-2D, and (l) SMGAN-3D. The red rectangle indicates the region zoomed in Fig.~\ref{fig: example3_roi}. This display window is [-160, 240]HU.}
\label{fig: example3}
\end{figure*}

\begin{figure}[!ht]
\centering
\subfloat[Full Dose FBP]{\includegraphics[width=0.80in]{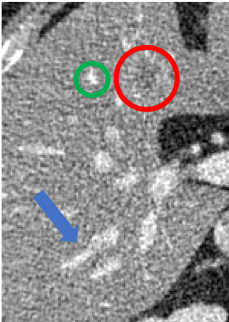}\label{fig: full2_roi}}\
\subfloat[Quarter Dose FBP]{\includegraphics[width=0.80in]{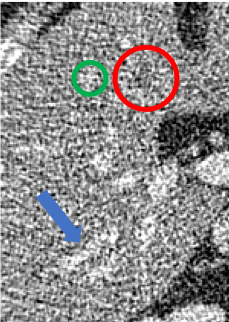}\label{fig: quarter2_roi}}\
\subfloat[CNN-L2]{\includegraphics[width=0.80in]{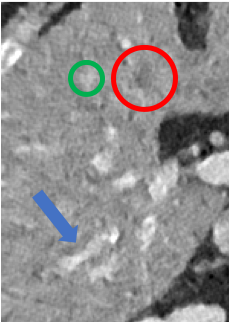}\label{fig: mse2_roi}}\
\subfloat[CNN-L1]{\includegraphics[width=0.80in]{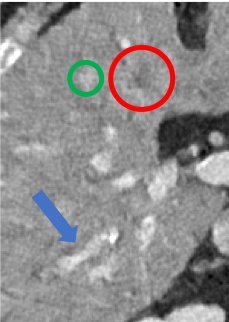}\label{fig: rmse2_roi}}

\subfloat[CNN-SL]{\includegraphics[width=0.80in]{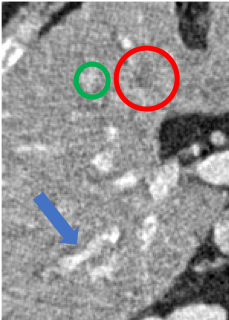}\label{fig: sl2_roi}}\
\subfloat[CNN-MSL]{\includegraphics[width=0.80in]{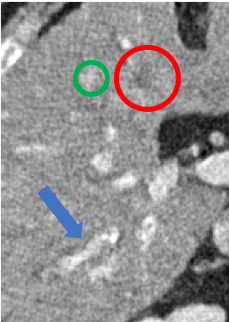}\label{fig: msl2_roi}}\
\subfloat[WGAN]{\includegraphics[width=0.80in]{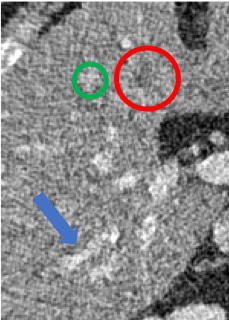}\label{fig: wgan2_roi}}\
\subfloat[BM3D]{\includegraphics[width=0.80in]{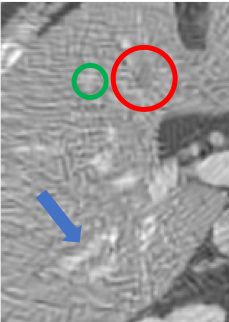}\label{fig: bm3d2_roi}}

\subfloat[RED-CNN]{\includegraphics[width=0.80in]{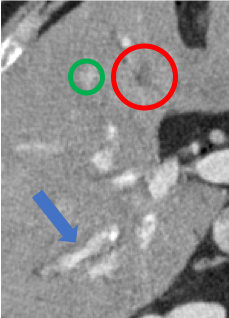}\label{fig: red_cnn2_roi}}\
\subfloat[WGAN-VGG]{\includegraphics[width=0.80in]{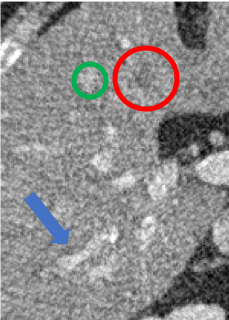}\label{fig: wgan_vgg2_roi}}\
\subfloat[SMGAN-2D]{\includegraphics[width=0.80in]{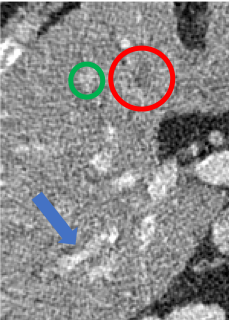}\label{fig: lsagan_2d2_roi}}\
\subfloat[SMGAN-3D]{\includegraphics[width=0.80in]{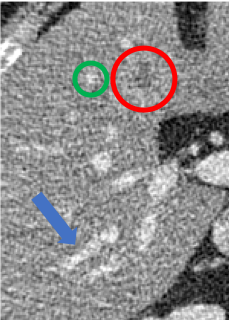}\label{fig: lsagan_3d2_roi}}
\caption{Zoomed parts of the region of interests~(ROIs) marked by the red rectangle in Fig.~\ref{fig: example3}. (a) NDCT, (b) LDCT, (c) CNN-L2, (d) CNN-L1, (e) SL-net, (f) MSL-net, (g) WGAN, (h) BM3D, (i) RED-CNN, (j) WGAN-VGG, (k) SMGAN-2D and (l) SMGAN-3D. The red circle indicates the metastasis and the green and blue arrows indicates two subtle structures. The display window is [-160,240]HU.}
\label{fig: example3_roi}
\end{figure}

In Fig.~\ref{fig: rmse_loss_convergence} and~\ref{fig: structural_loss_convergence}, in terms of $L_1$ and SL, we observe that $L_1$-net and $L_2$-net achieved the fastest convergence rate and have similar convergence trends in that all curves decreased initially and then smoothly converged, indicating that these mean-based algorithms both have fast convergence rates. Fig.~\ref{fig: rmse_loss_convergence} shows that they both converged around the $6^{th}$ epoch. In contrast, in Fig.~\ref{fig: rmse_loss_convergence}, there are differences between SL-based and mean-based methods. We can see that the convergence curve of the SL-net decreases initially and then slightly rises around the $4^{th}$ epoch as shown in Fig.~\ref{fig: rmse_loss_convergence}. MSL-net also shows a small increase like SL-net in terms of $L_1$. This observation indicates that SL-based and mean-based methods have different emphasis on minimizing perceptually motivated similarity between real NDCT images and generated NDCT images. For WGAN-based methods, it can be clearly observed that the curves for WGAN, WGAN-VGG, SMGAN-2D, and SMGAN-3D slightly oscillate in the convergence process after the $5^{th}$ epoch in Fig.~\ref{fig: rmse_loss_convergence} and~\ref{fig: structural_loss_convergence}. The reason for such oscillatory behaviors is as follows: $G$ attempts to mimic the real NDCT distribution while $D$ aims to differentiate between the real NDCT distribution and the denoised LDCT distribution. Since GAN's intrinsic nature is a two-player game, the distributions of $G$ and $D$ are constantly changing, and this leads to the oscillatory behavior when converging to their optimal status.

As shown in Fig.~\ref{fig: wgan_loss_convergence}, we can evaluate the convergence performance of WGAN. It can be seen that our proposed SMGAN-2D has the mildest oscillatory behavior compared with the other three models and reaches a stable state after the $13^{th}$ epoch. Moreover, the SMGAN-3D oscillates in a relatively large range in the training process. This is because our proposed SMGAN-3D considers 3D structural information which results in a relatively larger vibrating amplitude in the training process. However, the curve still oscillates close to the x-axis, indicating SMGAN-3D's robustness in minimizing the Wasserstein distance between the generated samples and real samples.

\subsection{Denoising Performance}
\label{subsec: denoising_performance}

To demonstrate the effectiveness of the proposed network, we perform the qualitative comparisons over three representative abdominal images presented in Figs.~\ref{fig: example1},~\ref{fig: example3} and~\ref{fig: example4}. For better evaluations of the image quality with different denoising models, zoomed regions-of-interest (ROIs) are marked by red rectangles and shown in Figs.~\ref{fig: example1_roi},~\ref{fig: example3_roi} and~\ref{fig: example4_roi} respectively. Note that all results from different denoising models focus on two aspects: content restoration and noise-reduction. All CT images in axial view are displayed in the angiography window [-160, 240]HU.

The real NDCT images and corresponding LDCT images are presented in Figs.~\ref{fig: full1}~and~\ref{fig: quarter1}. As observed, there are distinctions between ground truth (NDCT) images and LDCT images. Figs.~\ref{fig: full1} and~\ref{fig: full3} show the lesions/metastasis. Fig.~\ref{fig: full2} presents focal fatty sparing/focal fat. In Figs.~\ref{fig: full1_roi},~\ref{fig: full2_roi} and~\ref{fig: full3_roi}, these lesions can be clearly observed in NDCT images; in contrast, from Figs.~\ref{fig: quarter1_roi},~\ref{fig: quarter2_roi}, and~\ref{fig: quarter3_roi}, it can be seen that the original LDCT image is noisy, and lacks structural features for task-based clinical diagnosis. All adopted denoising models suppress noise to some extent.

\subsubsection{\textbf{Comparison with CNN-based denoising methods}}
\label{subsec:comparison_cnn}
To study the robustness of the adversarial learning framework in SMGAN-3D, we compared SMGAN-3D with the CNN-based methods, including CNN-L2, CNN-L1, RED-CNN~\cite{chen_zhang_kalra_lin_chen_liao_zhou_wang_2017}, SL-net and MSL-net. It is worth noting that CNN-L2, CNN-L1, and RED-CNN are mean-based denoising methods, and SL-net and MSL-net are SL-based denoising methods. All of the methods greatly reduce the noise compared with LDCT images. Our proposed method preserves more structural details, thereby yielding better image quality, compared with the other five methods.

Mean-based methods can effectively reduce noise, but the side effect is impaired image contents. In Fig.~\ref{fig: mse1}, $L_2$-net greatly suppresses the noise, but blurs some crucial structural information in the porta hepatis region. Meanwhile, some waxy artifacts can still be observed in Fig.~\ref{fig: mse2_roi}. $L_2$-net does not produce good visual quality because it assumes that the noise is independent of local characteristics of the images. Even though it retains high SNR, its results are not clinically preferable. Compared with $L_2$-net, in Figs.~\ref{fig: rmse1} and~\ref{fig: rmse2}, it can been seen that $L_1$-net encourages less blurring and preserves more structural information. However, as observed in Fig.~\ref{fig: rmse1_roi}, it still over-smooths some anatomical details. Meanwhile, in Fig.~\ref{fig: rmse2_roi}, there are some blocky effects marked by the blue arrow. The results obtained by RED-CNN~\cite{chen_zhang_kalra_lin_chen_liao_zhou_wang_2017} deliver high SNR but blur the vessel details as shown in Figs.~\ref{fig: red_cnn1_roi} and~\ref{fig: red_cnn2_roi}. 

For SL-based methods, as observed in Figs.~\ref{fig: sl1} and~\ref{fig: sl2}, SL-net generates images with higher contrast resolution and preserves texture of real NDCT images better than $L_2$-net and $L_1$-net. However, Figs.~\ref{fig: sl1_roi} and~\ref{fig: sl2_roi} show that SL-net does not preserve the structural features well, and there still remain small streak artifacts. Subsequently, in Figs.~\ref{fig: sl1_roi} and \ref{fig: msl1_roi}, SL-net and MSL-net have low frequency image intensity variance because SSIM/MS-SSIM is insensitive to uniform biases~\cite{wang2004image,wang2003multiscale}. On the other hand, $L_1$-net preserves the overall image intensity, but it does not preserve high contrast resolution well as SL-net and MSL-net do.

From Figs.~\ref{fig: example4} and~\ref{fig: example4_roi}, we can see mean-based and SL-based methods work well with effective noise suppression and artifact removal. However, the illustrations in Fig.~\ref{fig: example4_roi} show that these methods blur the local strutural features. Our proposed SMGAN-based methods present a better edge preservation than the competing methods.

Overall, the observations above support the following statements. First, although the voxel-wise methods show good noise-reduction properties, to some extent they blur the contents and lead to the loss of structural details because they optimize the results in the voxel-wise manner. Second, SL-based methods better preserve texture than mean-based methods, but they cannot preserve overall image intensity. Third, the results produced by the proposed SMGAN-3D demonstrate the benefits of the combination of two loss functions and the importance of the adversarial training~\cite{goodfellow2014generative,arjovsky2017wasserstein}.

\begin{table*}[!t]
\renewcommand{\arraystretch}{1.3}
\centering
\caption{Quantitative results associated with different approaches in Figs.~\ref{fig: example1} and~\ref{fig: example3}.}
\setlength{\tabcolsep}{5pt}
\begin{tabular}{c c c c c c c c c c c c}
\hline
& \multicolumn{3}{c}{Fig.~\ref{fig: example1}} && \multicolumn{3}{c}{Fig.~\ref{fig: example3}} && \multicolumn{3}{c}{Fig.~\ref{fig: example4}}\\
& PSNR & SSIM & RMSE && PSNR & SSIM & RMSE && PSNR & SSIM & RMSE \\
\cline{2-4}\cline{6-8}\cline{10-12}
LDCT & 22.818 & 0.761 & 0.0723 && 21.558 & 0.659 & 0.0836 && 24.169 & 0.737 & 0.0618 \\
CNN-L1 & 27.791 & 0.822 & 0.0408 && 26.794 & 0.738 & 0.0457 && 29.162 & 0.807 & 0.0348 \\
CNN-L2 & 27.592 & 0.819 & 0.0418 && 26.630 & 0.736 & 0.0466 && 28.992 & 0.806 & 0.0355 \\
SL-net & 26.864 & \textbf{0.831} & 0.0453 && 25.943 & \textbf{0.745} & 0.0504 && 28.069 & \textbf{0.813} & 0.0395 \\
MSL-net & 27.667 & \textbf{0.831} & 0.0414 && 26.685 & 0.744 & 0.0469 && 28.902 & 0.812	&0.0359 \\
WGAN & 25.727 & 0.801 & 0.0517 && 24.655 & 0.711 & 0.0585 && 26.782 & 0.781 & 0.0458 \\
BM3D & 27.312 & 0.809 & 0.0431 && 26.525 & 0.728 & 0.0472 && 28.959 & 0.794 & 0.0356 \\
RED-CNN & \textbf{28.279} & 0.825 & \textbf{0.0385} && \textbf{27.243} & 0.743 & \textbf{0.0444} && \textbf{29.679} & 0.811 & \textbf{0.0328} \\
WGAN-VGG & 26.464 & 0.811 & 0.0475 && 25.300 & 0.722 & 0.0543 && 27.161 & 0.793 & 0.0419\\
SMGAN-2D & 26.627 & 0.821 & 0.0466 && 25.507 & 0.732 & 0.0530 && 27.731 & 0.795 & 0.0406\\
SMGAN-3D & 26.569 & 0.824 & 0.0473 && 25.372 & 0.739 & 0.0538 && 27.398 & 0.794 & 0.0411\\
\hline
\end{tabular}
\label{table: psnr&ssim&rmse}
\end{table*}

\subsubsection{\textbf{Comparison with WGAN-based denoising methods}}
\label{subsec:comparison_wgan}
To evaluate the effectiveness of our proposed objective function, we compare our method with existing WGAN-based networks, including WGAN and WGAN-VGG. Considering the importance of clinical image quality and specific structural features for medical diagnosis, we adopted the adversarial learning method~\cite{goodfellow2014generative,arjovsky2017wasserstein} in our experiments because WGAN could help to capture more structural information. 
Nevertheless, based on our prior experience, utilizing WGAN alone may yield stronger noise than other selected approaches, because it only maps the data distribution from LDCT to NDCT without consideration of local voxel intensity and structural correlations. The observations demonstrate that the noise texture is coarse in the images, as shown in Fig.~\ref{fig: wgan1_roi} and Fig.~\ref{fig: wgan3_roi}, which support our intuition. 

Indeed, the images of WGAN-VGG\cite{yang2017low}, as shown in Fig.~\ref{fig: wgan_vgg1}, exhibit better visual quality with respect to more details and share structural details similar to NDCT images according to human perceptual evaluations. However, Figs.~\ref{fig: wgan_vgg1_roi}~(marked by the red circle) and~\ref{fig: wgan_vgg2_roi}~(marked by the green circle) suggest that it may severely distort the original structural information. A possible reason is that the VGG network ~\cite{simonyan2014very} is a pre-trained deep CNN network based on natural images, and the structural information and contents of natural images are different from medical images.

Compared with WGAN and WGAN-VGG, our proposed SMGAN-3D, as shown in Figs.~\ref{fig: lsagan_3d1_roi}~(marked by the red circle) and~\ref{fig: lsagan_3d2_roi}~(marked by the green circle), can more clearly visualize the metastasis and better preserve of the portal vein.

\begin{figure*}[!t]
\centering
\subfloat[Full Dose FBP]{\includegraphics[width=1.7in]{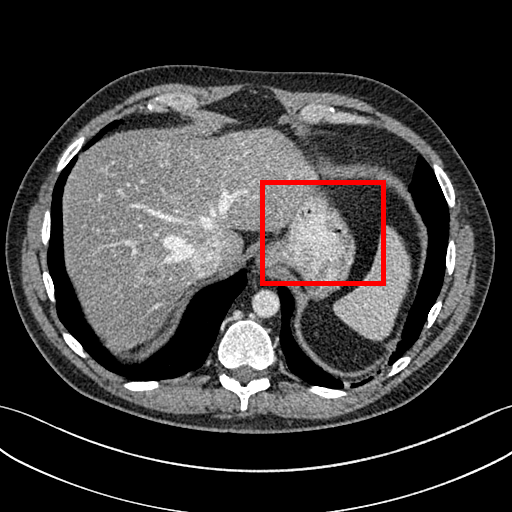}\label{fig: full3}}
\subfloat[Quarter Dose FBP]{\includegraphics[width=1.7in]{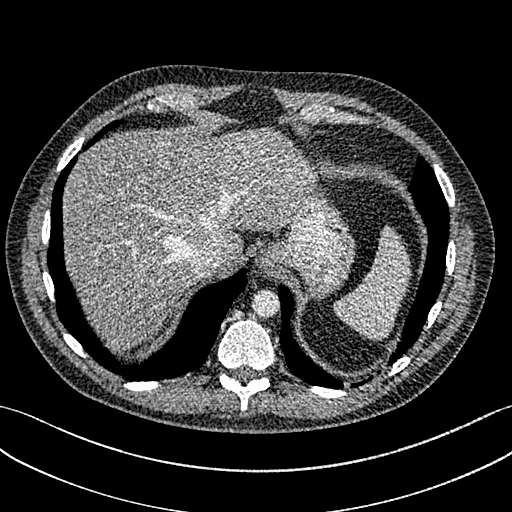}\label{fig: quarter3}}
\subfloat[CNN-L2]{\includegraphics[width=1.7in]{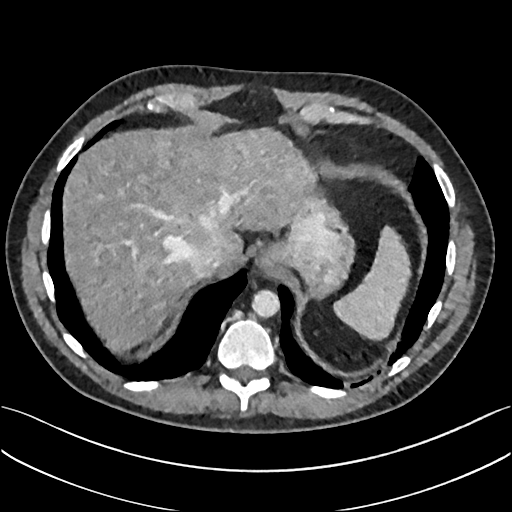}\label{fig: mse3}}
\subfloat[CNN-L1]{\includegraphics[width=1.7in]{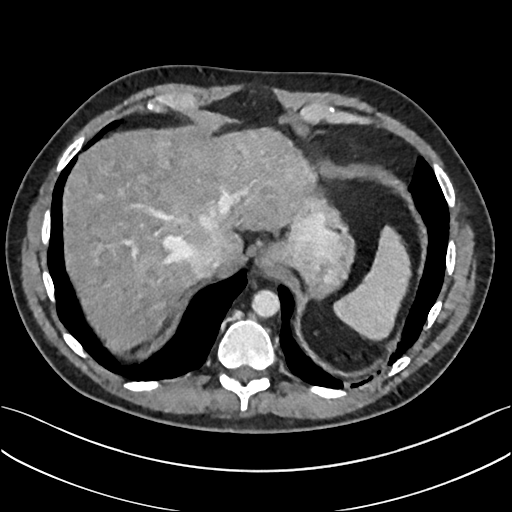}\label{fig: rmse3}}

\subfloat[CNN-SL]{\includegraphics[width=1.7in]{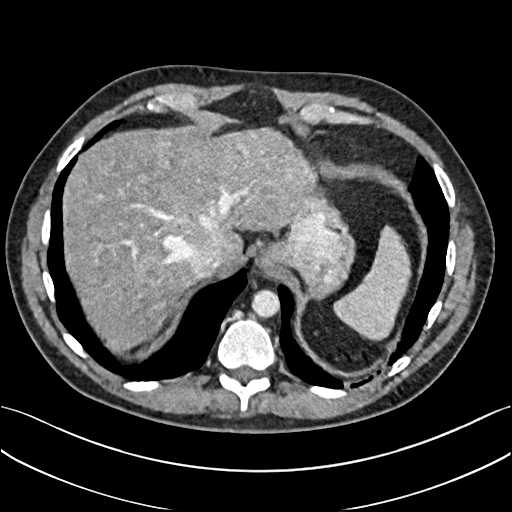}\label{fig: sl3}}
\subfloat[CNN-MSL]{\includegraphics[width=1.7in]{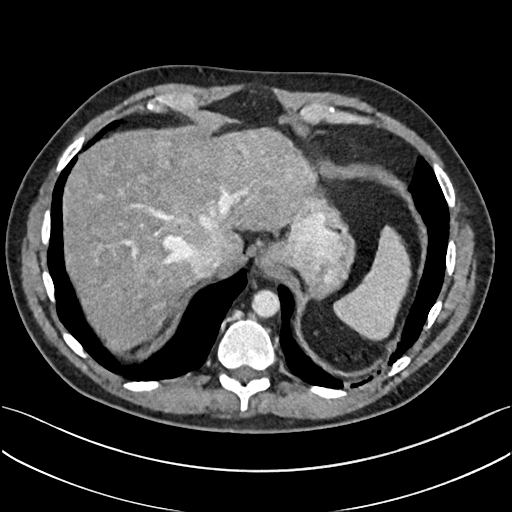}\label{fig: msl3}}
\subfloat[WGAN]{\includegraphics[width=1.7in]{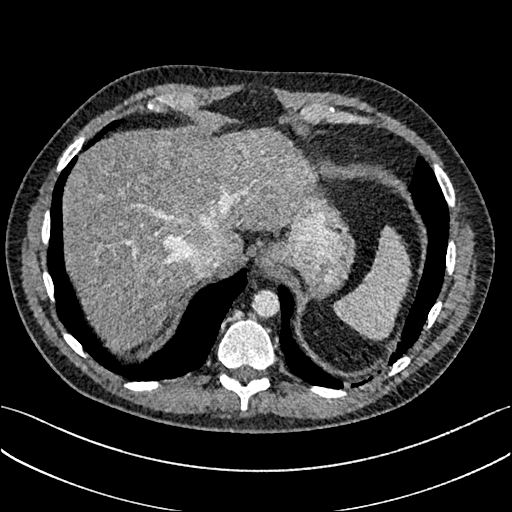}\label{fig: wgan3}}
\subfloat[BM3D]{\includegraphics[width=1.7in]{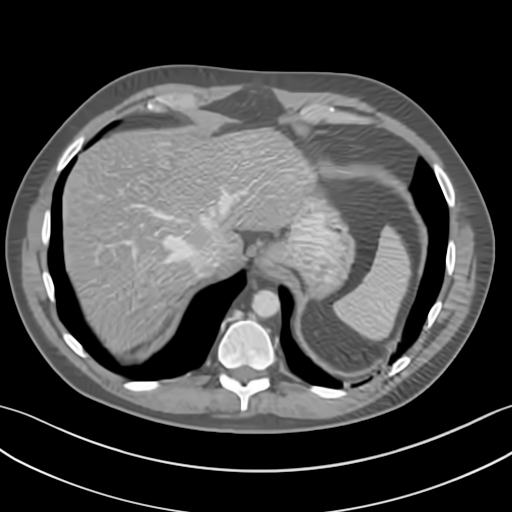}\label{fig: bm3d3}}

\subfloat[RED-CNN]{\includegraphics[width=1.7in]{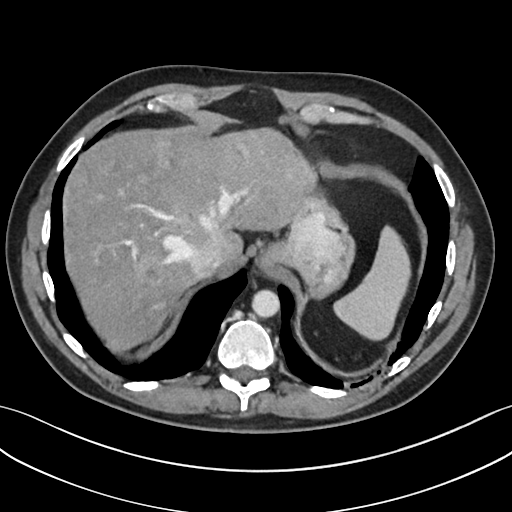}\label{fig: red_cnn3}}
\subfloat[WGAN-VGG]{\includegraphics[width=1.7in]{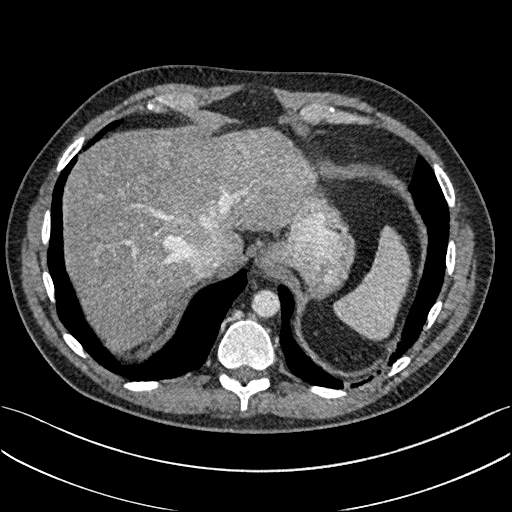}\label{fig: wgan_vgg3}}
\subfloat[SMGAN-2D]{\includegraphics[width=1.7in]{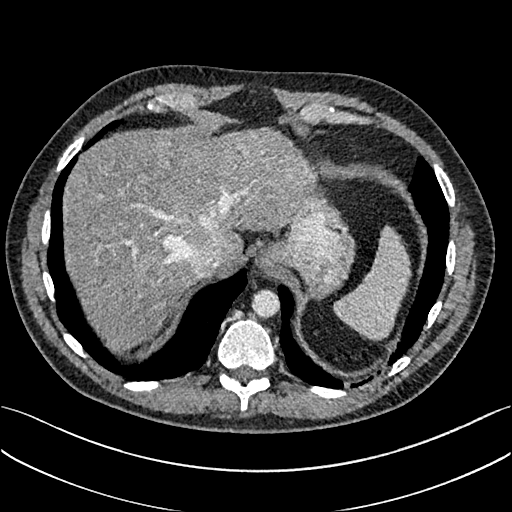}\label{fig: lsagan_2d3}}
\subfloat[SMGAN-3D]{\includegraphics[width=1.7in]{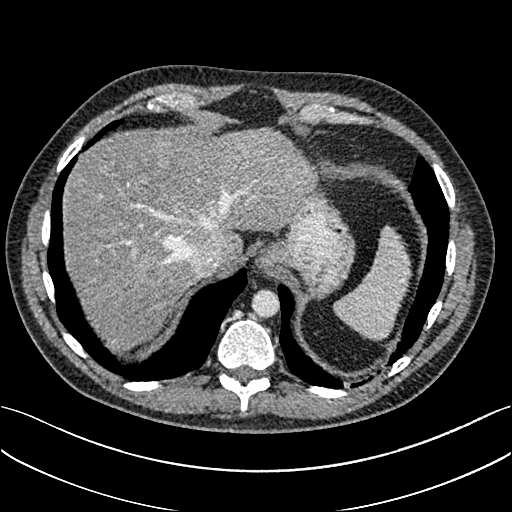}\label{fig: lsagan_3d3}}
\caption{Results from abdomen CT images.~(a) NDCT, (b) LDCT, (c) CNN-L2, (d) CNN-L1, (e) SL-net, (f) MSL-net, (g) WGAN (h) BM3D, (i) RED-CNN, (j) WGAN-VGG, (k) SMGAN-2D, and (l) SMGAN-3D. The red rectangle indicates the region zoomed in Fig.~\ref{fig: example4_roi}. This display window is [-160, 240]HU.}
\label{fig: example4}
\end{figure*}

\begin{figure}[!t]
\centering
\subfloat[Full Dose FBP]{\includegraphics[width=0.82in]{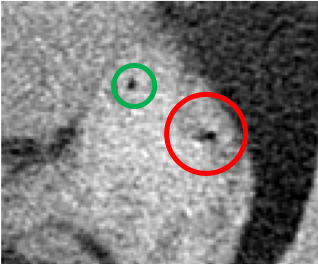}\label{fig: full3_roi}}\
\subfloat[Quarter Dose FBP]{\includegraphics[width=0.82in]{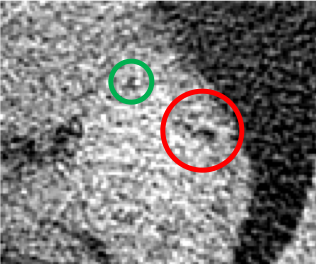}\label{fig: quarter3_roi}}\
\subfloat[CNN-L2]{\includegraphics[width=0.82in]{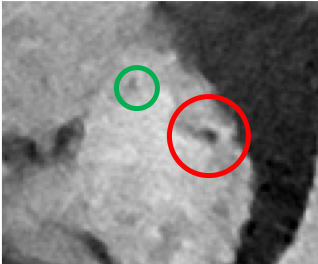}\label{fig: mse3_roi}}\
\subfloat[CNN-L1]{\includegraphics[width=0.82in]{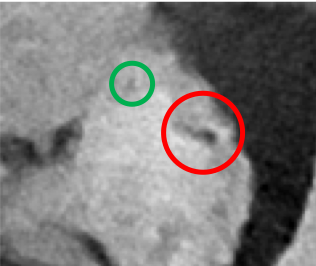}\label{fig: rmse3_roi}}

\subfloat[CNN-SL]{\includegraphics[width=0.82in]{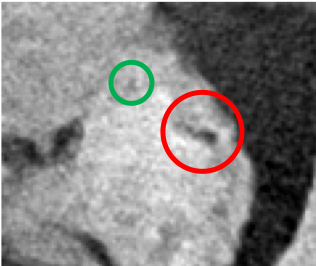}\label{fig: sl3_roi}}\
\subfloat[CNN-MSL]{\includegraphics[width=0.82in]{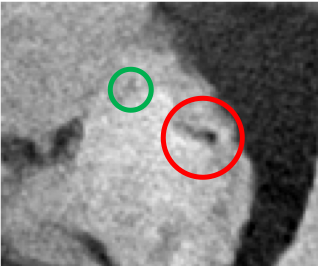}\label{fig: msl3_roi}}\
\subfloat[WGAN]{\includegraphics[width=0.82in]{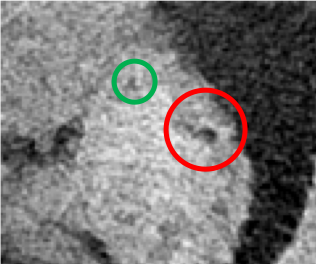}\label{fig: wgan3_roi}}\
\subfloat[BM3D]{\includegraphics[width=0.82in]{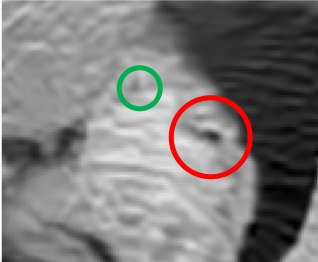}\label{fig: bm3d3_roi}}

\subfloat[RED-CNN]{\includegraphics[width=0.82in]{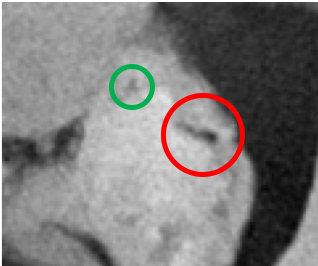}\label{fig: red_cnn3_roi}}\
\subfloat[WGAN-VGG]{\includegraphics[width=0.82in]{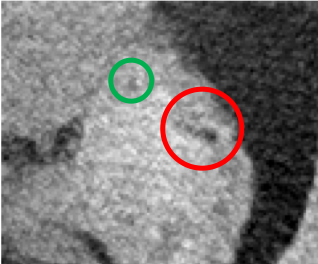}\label{fig: wgan_vgg3_roi}}\
\subfloat[SMGAN-2D]{\includegraphics[width=0.82in]{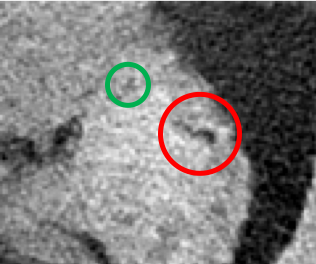}\label{fig: lsagan_2d3_roi}}\
\subfloat[SMGAN-3D]{\includegraphics[width=0.82in]{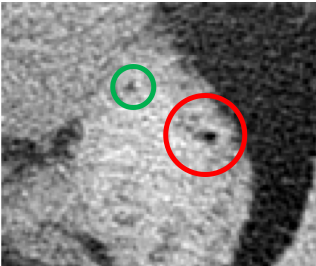}\label{fig: lsagan_3d3_roi}}
\caption{Zoomed parts of the region of interests~(ROIs) marked by the red rectangle in Fig.~\ref{fig: example4}. (a) NDCT, (b) LDCT, (c) CNN-L2, (d) CNN-L1, (e) SL-net, (f) MSL-net, (g) WGAN, (h) BM3D, (i) RED-CNN, (j) WGAN-VGG, (k) SMGAN-2D and (l) SMGAN-3D. The red and the green circles indicate subtle edges. The display window is [-160,240]HU.}
\label{fig: example4_roi}
\end{figure}

In Figs.~\ref{fig: example4} and~\ref{fig: example4_roi}, it can be found that the SMGAN-based methods can achieve better anatomical feature preservations and visual quality than other state-of-the-art methods.

The experimental results demonstrate that our proposed objective function is essential to capture more accurate anatomical details.

\subsubsection{\textbf{Comparison with Image space denoising}}
\label{subsec:comparison_exist_ldct}
To validate the robustness of DL-based methods, we compared our method with the image space denoising method. Figs.~\ref{fig: bm3d1_roi} and~\ref{fig: bm3d2_roi} show that BM3D blurs the low-contrast lesion marked by the red circle and smooths specific features marked by the blue arrow. In contrast, SMGAN-3D exhibits better on the low-contrast lesion and yields sharper features as shown in Figs.~\ref{fig: lsagan_3d1_roi} and~\ref{fig: lsagan_3d2_roi}.

\subsubsection{\textbf{Comparison with 2D-based SMGAN network}}
\label{subsec:comparison_2d}
In order to evaluate the 3D structural information, we compared SMGAN-3D with SMGAN-2D. As shown in Fig.~\ref{fig: lsagan_3d1_roi}, our proposed SMGAN-3D generated the results with better subtle details than SMGAN-2D and enjoys more similar statistical noise properties to the corresponding NDCT images. The reasons why SMGAN-3D outperforms SMGAN-2D are follows. First, SMGAN-3D incorporates 3D structural information to improve image quality. Second, SMGAN-2D takes input slice by slice, thus potentially leading to the loss of spatial correlation between adjacent slices. 

Figs.~\ref{fig: example4} and~\ref{fig: example4_roi} demonstrate that the SMGAN-3D can be used to provide improved anatomical feature preservation over other state-of-the-art methods.

In summary, we compared our proposed methods with existing methods, and it can be clearly observed that SMGAN-3D achieves robust performance in noise suppression, artifact removal, and texture preservation. Note that we recommend the reader to see ROIs (in Fig.~\ref{fig: example1_roi} and~\ref{fig: example3_roi}) or zoom in to better evaluate our results. To further validate the generalization ability of our proposed model, we conclude more details in Appendix~\ref{subsec:differenttrainingsets}.

\begin{figure}[t]
\centering
\subfloat[]{\includegraphics[width=3in]{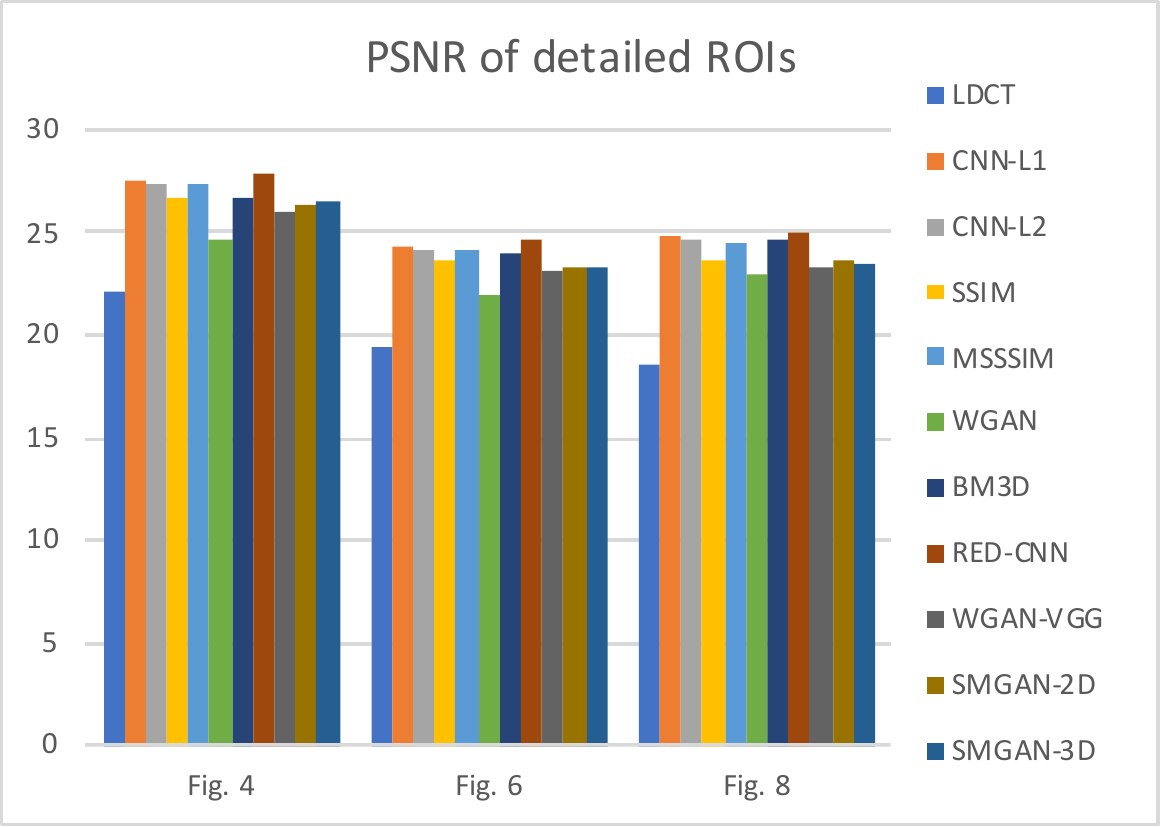}\label{fig: psnr_roi}}

\subfloat[]{\includegraphics[width=3in]{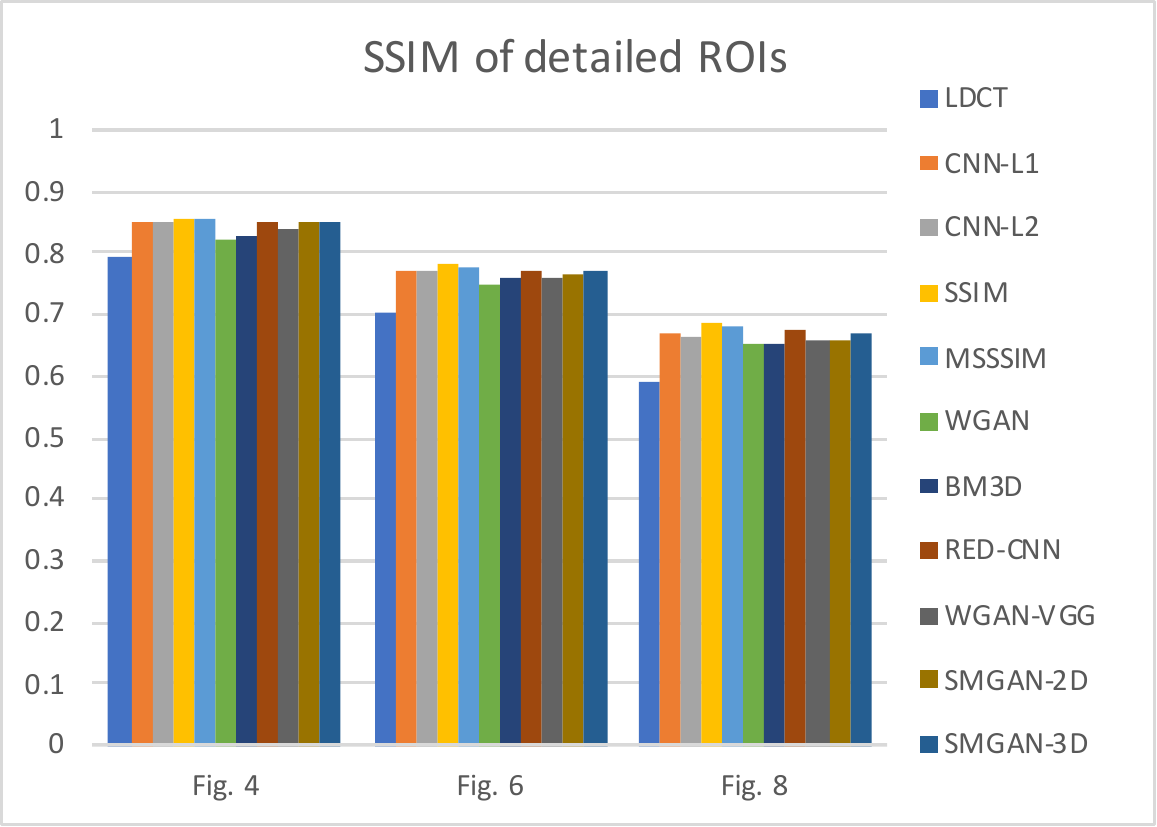}\label{fig: ssim_roi}}

\subfloat[]{\includegraphics[width=3in]{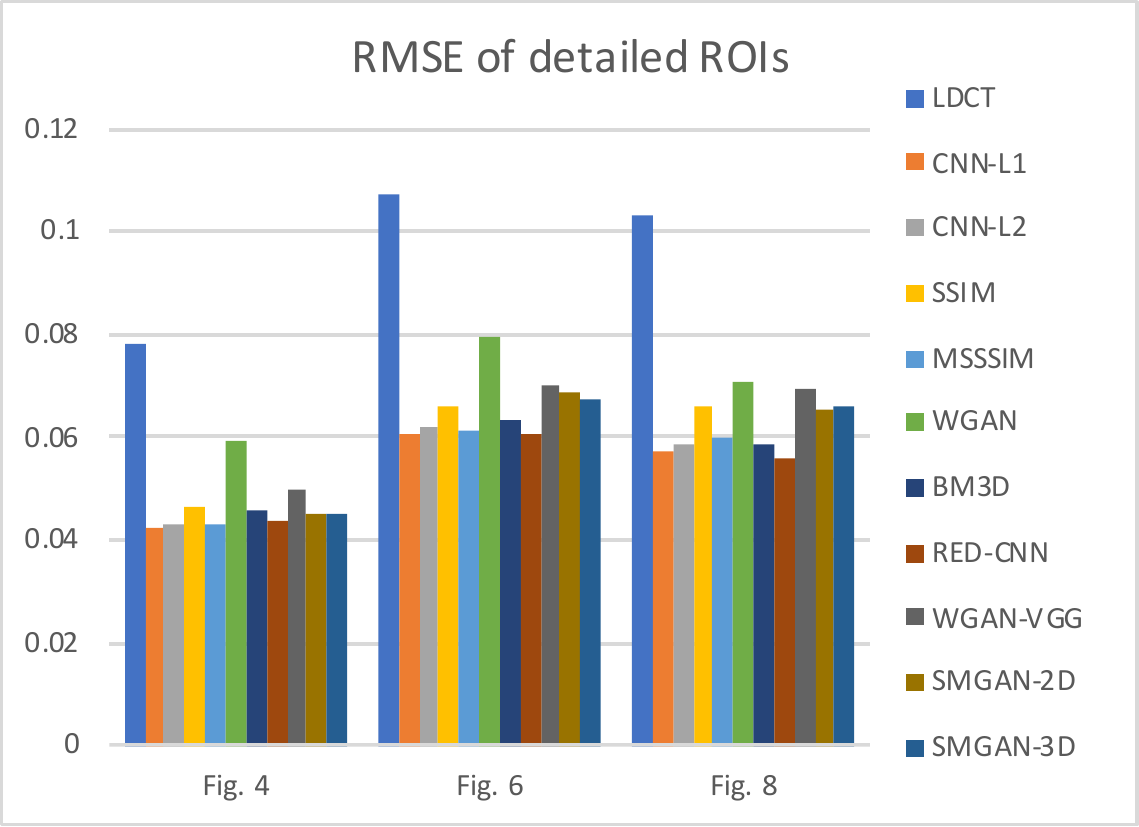}\label{fig: rmse_roi}}

\caption{Performance comparison of LDCT and ten algorithms over the ROIs marked by the red rectangles in Fig.~\ref{fig: full1} and Fig.~\ref{fig: full2}.} 
\label{fig: roi_metrics}
\end{figure}

\begin{table*}[!t]
\renewcommand{\arraystretch}{1.3}
\centering
\caption{Statistical properties of the images in Figs.~\ref{fig: example1_roi},~\ref{fig: example3_roi} and ~\ref{fig: example4_roi}. These are the ROIs indicated by the red rectangles in Figs.~\ref{fig: example1},~\ref{fig: example3} and~\ref{fig: example4}. Note that the relative percentage difference of NDCT values versus the rest of models is added to aid the readers.}
\begin{tabular}{c c c c c c c c c}
\hline
& \multicolumn{2}{c}{Fig.~\ref{fig: example1_roi}} && \multicolumn{2}{c}{Fig.~\ref{fig: example3_roi}} && \multicolumn{2}{c}{Fig.~\ref{fig: example4_roi}} \\
& Mean & SD && Mean & SD && Mean & SD \\
\cline{2-3}\cline{5-6}\cline{8-9}
NDCT & 115.282 & 45.946 && 56.903 & 58.512 && 51.225 & 73.297\\
LDCT & 114.955 (-0.2837\%) & \textbf{74.299 (61.709\%)} && 57.228 (0.571\%) & \textbf{85.854 (46.729\%)} && 50.142 (-2.114\%) & 89.346 (21.896\%)\\
CNN-L1 & 115.809 (0.4571\%) & 28.532 (-37.9010\%) && 57.709 (1.416\%) & 42.315 (-27.682\%) && 50.917 (-0.6013\%) & 66.359 (-9.466\%)\\
CNN-L2 & 117.191 (1.656\%) & 29.933 (-34.852\%) && 58.956 (3.608\%) & 43.411 (-25.808\%) && 52.229 (1.960\%) & 66.922 (-8.698\%)\\
SL-net & \textbf{131.333 (13.923\%)} & 35.844 (-21.987\%) && 68.471 (20.329\%) & 50.789 (-13.199\%) && 63.874 (24.693\%) & 72.718 (-0.790\%)\\
MSL-net & 118.395 (2.701\%) & 32.548 (-29.160\%) && 63.271 (11.191\%) & 46.979 (-19.711\%) && 57.052 (11.375\%) & 69.519 (-5.154\%)\\
WGAN & 105.461 (-8.519\%) & 42.659 (-7.154\%) && 48.432 (-14.887\%) & 54.306 (-7.188\%) && 42.417 (-17.195\%) & 70.904 (-3.265\%)\\
BM3D & 114.058 (-1.062\%) & 31.515 (-31.409\%) && \textbf{25.649 (-54.925\%)} & 69.411 (18.627\%) && \textbf{15.183 (-70.360\%)} & \textbf{100.08 (36.540\%)}\\
RED-CNN & 116.642 (1.180\%) & 27.194 (-40.813\%) && 57.985 (1.902\%) & 42.048 (-28.138\%) && 51.272 (0.0918\%) & 66.961 (-8.644\%)\\
WGAN-VGG & 108.229 (-6.118\%) & 36.721 (-20.078\%) && 54.450 (-4.311\%) & 48.660 (-16.838\%) && 44.959 (-12.232\%) & 67.059 (-8.511\%)\\
SMGAN-2D & 108.758 (-5.659\%) & 40.948 (-10.878\%) && 51.243 (-9.947\%) & 53.065 (-9.309\%) && 48.230 (-5.847\%) & 72.073 (-1.670\%)\\
SMGAN-3D & 115.569 (0.749\%) & 43.654 (-6.723\%) && 54.356 (-4.476\%) & 56.552 (-3.350\%) && 55.378 (8.107\%) & 73.303 (-0.00821\%)\\
\hline
\end{tabular}
\label{table: stats}
\end{table*}

\begin{table*}[!t]
\renewcommand{\arraystretch}{1.5}
\centering
\caption{Visual assessment scores by three radiologist readers.}
\setlength{\tabcolsep}{10pt}
\begin{tabular}{c c c c c c}
\hline
& Sharpness & Noise Suppression & Diagnostic Acceptability & Contrast Retention & Overall Quality \\
\cline{2-6}
LDCT 		& 2.55$\pm$1.43 			& 1.55$\pm$0.80 	& 1.85$\pm$0.96 			& 1.75$\pm$0.83 			& 1.93$\pm$1.01 \\
CNN-L1 		& 2.80$\pm$0.81 			& 3.30$\pm$0.71 	& 2.70$\pm$0.78 			& 2.75$\pm$0.77 			& 2.89$\pm$0.77 \\
CNN-L2 		& 2.12$\pm$0.42 			& \textbf{3.98$\bm{\pm}$0.58} & 1.93$\pm$0.78 	& 2.07$\pm$0.83 			& 2.53$\pm$0.55 \\
SL-net 		& 2.95$\pm$0.86 			& 3.15$\pm$0.65 	& 2.70$\pm$0.71 			& 2.80$\pm$0.81 			& 2.90$\pm$0.76 \\
MSL-net 		& 3.01$\pm$0.94 			& 3.16$\pm$0.57 	& 2.87$\pm$0.83 			& 2.84$\pm$0.69 			& 2.97$\pm$0.76 \\
WGAN 		& 3.30$\pm$0.56 			& 2.80$\pm$0.81 	& 3.15$\pm$0.91 			& 3.45$\pm$1.02 			& 3.09$\pm$0.66 \\
BM3D 		&  2.21$\pm$1.08 			& 3.29$\pm$0.80 	& 2.21$\pm$0.86 			& 2.29$\pm$0.88 			& 2.50$\pm$0.91 \\
RED-CNN 	&  3.29$\pm$0.88			&  3.79$\pm$0.70 	& 3.51$\pm$0.70 			& 3.46$\pm$1.12			&  3.51$\pm$0.85 \\
WGAN-VGG 	& 3.35$\pm$0.91 			& 3.50$\pm$1.07 	& 3.35$\pm$0.91 			& 3.45$\pm$1.02			& 3.41$\pm$0.94 \\
SMGAN-2D 	& 3.25$\pm$0.65			& 3.48$\pm$0.66	& 3.32$\pm$0.58			& 3.21$\pm$0.78			& 3.32$\pm$0.67 \\
SMGAN-3D 	& \textbf{3.56$\bm{\pm}$0.73}	& 3.59$\pm$0.68	& \textbf{3.58$\bm{\pm}$0.46}	& \textbf{3.61$\bm{\pm}$1.02}	& \textbf{3.59$\bm{\pm}$0.72} \\
\hline
\end{tabular}
\label{table: reader_study}
\end{table*}

\subsection{Quantitative analysis}
\label{subsec: quantitative_analysis}
We performed the quantitative analysis with respect to three selected metrics (PNSR, SSIM, and RMSE). Then, we investigated the statistical properties of the denoised images for each noise-reduction algorithm. Furthermore, we performed a blind reader study with three radiologists on 10 groups of images. Note that quantitative full-size measurements are in Table~\ref{table: psnr&ssim&rmse} and image quality assessments of ROIs are in Fig.~\ref{fig: roi_metrics}. The NDCT images are chosen as the gold-standard.

\subsubsection{\textbf{Image quality analysis}}
\label{subsec:comparison_image_quality}

As shown in Table~\ref{table: psnr&ssim&rmse}, RED-CNN scores the highest PSNR and RMSE, and ranks the second place in SSIM. Since the properties of PSNR and RMSE are regression to the mean, it is expected that RED-CNN, a mean-based regressiom optimization, has better performance than other feature-based models. For SL-net and MSL-net, it is not surprising that both models achieve the highest SSIM scores due to the adoption of structural similarity loss. However, a good score measured by image quality metrics does not ensure the preservation of high-level feature information and structural details, and this explains why RED-CNN can have the best PSNR and RMSE despite over-smoothing the content. PSNR, SSIM and RMSE are not perfect, and they are subject to image blurring abd blocky/waxy artifacts in the denoised images, as shown in Figs.~\ref{fig: example1}\,-\,\ref{fig: example4_roi}. Hence, these metrics may not be sufficient in evaluating image quality and indicating diagnostic performance. Indeed, WGAN can provide better visual quality and achieve improved statistical properties. Compared with the CNN-based methods, the WGAN architecture can progressively reserve the consistency of the feature distributions between LDCT and NDCT images. By encouraging less blurring, WGAN alone could introduce more image noise to compromise diagnosis. To keep information in LDCT images, our novel loss function with a regularization term is structurally alert to enhance the clinical usability as compared to the other methods.

Although mean-based approaches, such as $L_1$-net, $L_2$-net, enjoy high metric scores, they may over-smooth the overall image contents and lose feature characteristics, which do not satisfy our HVS requirements because mean-based methods favor the regression toward the mean. Meanwhile, WGAN-VGG satisfies HVS requirements, but gets the lowest scores in the three selected metrics. The reason for the lowest scores is that WGAN-VGG may suffer from loss of subtle structural information or noise features, which may severely affect the diagnostic accuracy. The proposed SMGAN-2D outperforms the feature-based method WGAN-VGG with reference to the three metrics, illustrating the robust denoising capability of our proposed loss function. Compared with the SMGAN-2D model, SMGAN-3D achieves higher scores in PSNR and SSIM since it incorporates 3D spatial information. To further validate the performance of each denoising model with respect to clinically significant local details, we performed the quantitative analysis over ROIs. The summary of the quantitative results from ROIs is shown in Fig.~\ref{fig: roi_metrics}. It is worth noting that the quantitative results of the ROIs follow a similar trend to that of the full-size images.

\subsubsection{\textbf{Statistical analysis}}
\label{subsec:comparison_Statistical}

To quantitatively evaluate the statistical properties of processed images by different denoising models, we calculate the mean CT number (Hounsfield Unit) and standard deviations (SDs) of ROIs, as shown in Table~\ref{table: stats}. For each denoising model, the percent error of the mean and SD values were calculated in comparison to those of the reference (NDCT) images. The lower percent errors correspond to more robust denoising models. As shown in Table~\ref{table: stats}, $L_1$-net, $L_2$-net, SL-net, MSL-net, BM3D, RED-CNN, and WGAN-VGG generate high percent errors in SD with respect to the NDCT images. There are blocky and over-smoothing effects in the images which match our visual inspections. Specifically, for Fig.~\ref{fig: example4_roi}, the absolute difference in SD between BM3D and NDCT is the largest among all of the denoising models, which indicates that BM3D has the most noticeable blurring effects. The standard deviation of BM3D supports our visual observations as shown in Figs.~\ref{fig: bm3d1_roi},~\ref{fig: bm3d2_roi}, and~\ref{fig: bm3d3_roi}. The mean values of WGAN, WGAN-VGG, SL-net and SMGAN-2D deviated much from that of the NDCT image in Fig.~\ref{fig: example1_roi}. This indicates that WGAN, WGAN-VGG, and SMGAN-2D effectively reduce the noise level but compromise significant content information. Nevertheless, the SD value of SMGAN-2D is close to that of NDCT, which indicates that it supports HVS requirements. From the quantitative analysis in Table~\ref{table: stats}, it can be observed that our proposed SMGAN-3D achieves the best matching SD to the NDCT images out of all other methods. Overall, SMGAN-3D is a highly competitive denoising model for clinical use.

\subsubsection{\textbf{Visual assessments}}
\label{subsec:visual_assessment}
To validate clinical image quality of processed results, three radiologists performed a visual assessment on 10 groups of images. Each group includes an original LDCT image with lesions, the corresponding reference NDCT image, and the processed images by different denoising methods. NDCT, considered as the gold-standard, is the only labeled image in each group. All other images were evaluated on sharpness, noise suppression, diagnostic acceptability, and contrast retention using a five-point scale (5 = excellent and 1 = unacceptable). We invited three radiologists with mean clinical experience of 12.3 years to join our study. Note that these results were evaluated independently and the overall image quality score for each method was computed an averaging score from the four evaluation criteria. For different methods, the final score is presented as mean$\,\pm\,$SD (average score of three radiologists$\,\pm\,$standard deviation). The final quantitative results are listed in Table~\ref{table: reader_study}. 

As observed, the original LDCT images have the lowest scores because of their severe image quality degradation. All denoising models improve the scores to some extent in this study. From Table~\ref{table: reader_study}, RED-CNN obtains the highest score in noise suppression. Compared to all other methods, our proposed SMGAN-3D scores best with respect to sharpness, diagnostic acceptability, and contrast retention. Furthermore, voxel-wise optimization (CNN-L2) has the best visually-assessed image noise suppression, but it suffers from relatively low scores in sharpness and diagnostic acceptability, indicating a loss of image details. The proposed SMGAN-3D model gets a superior overall image quality score relative to the 2D model, which indicates that a 3D model can enhance CT image denoising performance by incorporating spatial information from adjacent slices.

In brief, the visual assessment demonstrates that SMGAN-3D has powerful capabilities in noise reduction, subtle image structure and edge preservation, and artifact removal. Most importantly, it satisfies the HVS requirements as shown in Figs.~\ref{fig: example1}\,-\,\ref{fig: example3_roi}.

\subsection{Computational Cost}
In CT reconstruction, there is a trade-off between the computational cost and the image quality. In this aspect, a DL-based algorithm has great advantages in computational efficiency. Although the training of DL-based methods is time-consuming, it can rapidly perform the denoising tasks on reconstructed LDCT images after the training is completed. In our study, the proposed 2D method requires about 15 hours and the 3D model needs approximately 26 hours for training to converge. WGAN-VGG, which has the same number of layers, takes about 18 hours in the training phase. Compared with iterative reconstruction, any DL-based approach will require much less execution time, which facilitates the clinical workflow. In practice, our proposed SMGAN-2D and SMGAN-3D took 0.534s and 4.864s respectively in the validation phase on a NVIDA Titan GPU. Compared with the results in~\cite{liu2014gpu,matenine2015gpu}, our method took significantly less time. For example, the computational cost for soft threshold filtering (STF)-based TV minimization in the ordered-subset simultaneous algebraic reconstruction technique (OS-SART) framework took 45.1s per iteration on the same computing platform. Hence, it is clear that once the model is trained, it requires far less computational overhead than an iterative reconstruction method given that other conditions are equal.

\section{Discussions}
\label{sec:discussion}
As mentioned before, different emphases on visual evaluation and traditional image quality metrics were extensively investigated. When training with only the mean-based losses ($L_1$-net, $L_2$-net, RED-CNN), the results can achieve high scores in quantitative metrics and yield promising results with substantial noise reduction. When training with the feature-based methods (WGAN-VGG), the results can meet HVS requirements for visualization since they preserve more structural details than mean-based methods. However, these methods suffer from the potential risk of content distortion since a perceptual loss is computed based on a network~\cite{simonyan2014very} trained on a natural image dataset. Practically and theoretically, even though adversarial learning can prevent smoothing in the image, and capture structural characteristics, they may often result in severe loss of diagnostic information. To integrate the best characteristics of these loss functions, we have proposed a hybrid loss function to deliver the LDCT image quality optimally.

Although our proposed network has achieved high-quality denoised LDCT images, there are still rooms for potential improvements. First and foremost, some feature edges in the processed results still look blurry. Also, some structural variations between NDCT and LDCT do not perfectly match. A possible way to enhance correlation between NDCT and LDCT is to design a network with a better modeling capability, which is the work we have started. As far as our reader study is concerned, although visual assessment may be subject to intra- as well as inter-operator variability, on average such assessment can still evaluate different algorithms effectively, especially in a pilot study. In our follow-up study, we will invite more radiologists to rate the results, and then quantify inter-operator variability in a task-specific fashion, and also study intra-operator variability.

\section{Conclusion}
\label{sec:conclusions}
In conclusion, we have presented a 3D CNN-based method for LDCT noise reduction. As a follow-up to our previous work~\cite{yang2017low}, a 3D convolutional neural network is utilized to improve the image quality in the 3D contextual setting. In addition, we have highlighted that the purpose of loss functions is to preserve high-resolution and critical features for diagnosis. Different from the state-of-the-art LDCT denoising method used in~\cite{wolterink2017generative}, an efficient structurally-sensitive loss has been included to capture informative structural features. Moreover, we have employed the Wasserstein distance to stabilize the training process for GAN. We have performed the quantitative and qualitative comparison of the image quality. The assessments have demonstrated that SMGAN-3D can produce results with higher-level image quality for clinical usage compared with the existing denoising networks~\cite{chen2017low,chen_zhang_kalra_lin_chen_liao_zhou_wang_2017,wolterink2017generative,yang2017low}.

In the future, we will extend our model to other medical imaging modalities in a task-specific manner. Moreover, we plan to incorporate more advanced denoising models such as the networks mentioned in~\cite{dai2017deformable,he2016Deep,sabour2017dynamic} for LDCT reconstruction. Finally, we are also interested in making our denoising software robust over different scanners.

\appendices

\section{Different training sets for SMGAN-3D training}
\label{subsec:differenttrainingsets}
\begin{figure}[!ht]
\centering
\subfloat[CASE1]{\includegraphics[width=0.85in]{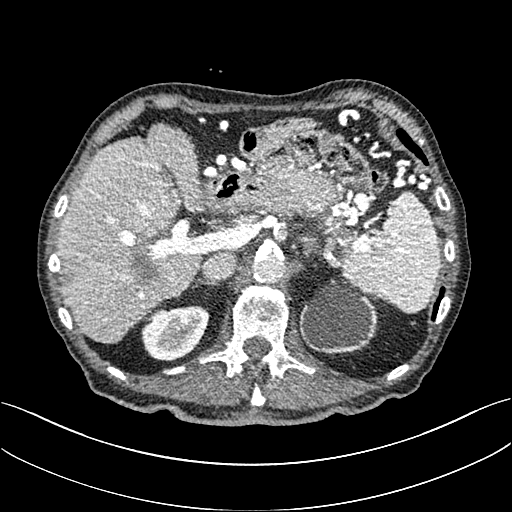}\label{fig: case11}}
\subfloat[CASE2]{\includegraphics[width=0.85in]{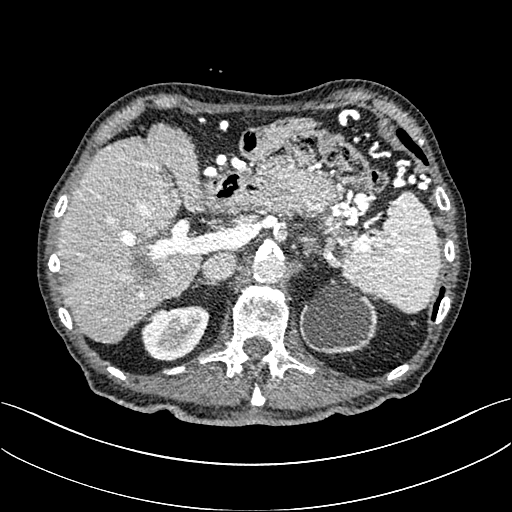}\label{fig: case12}}
\subfloat[CASE3]{\includegraphics[width=0.85in]{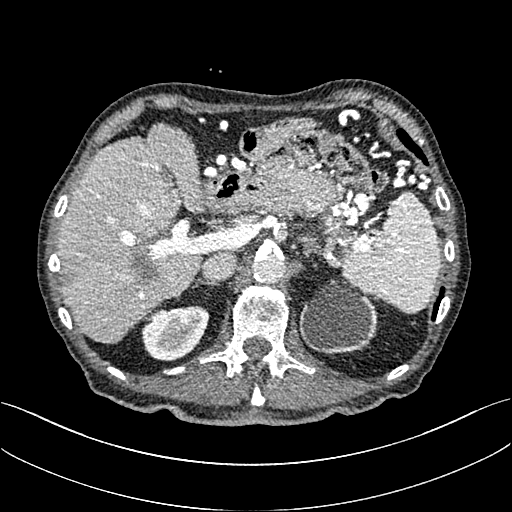}\label{fig: case13}}
\subfloat[CASE4]{\includegraphics[width=0.85in]{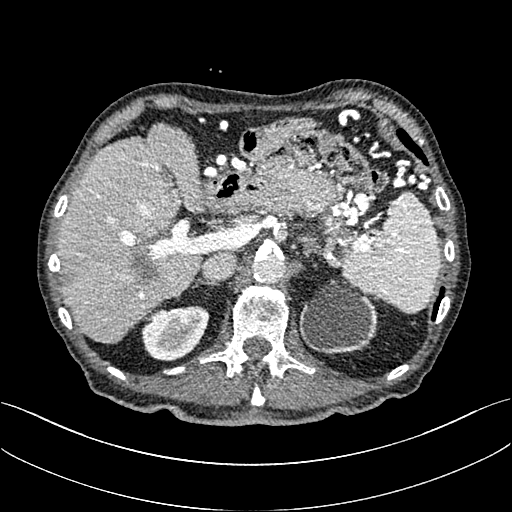}\label{fig: case14}}

\subfloat[CASE1]{\includegraphics[width=0.85in]{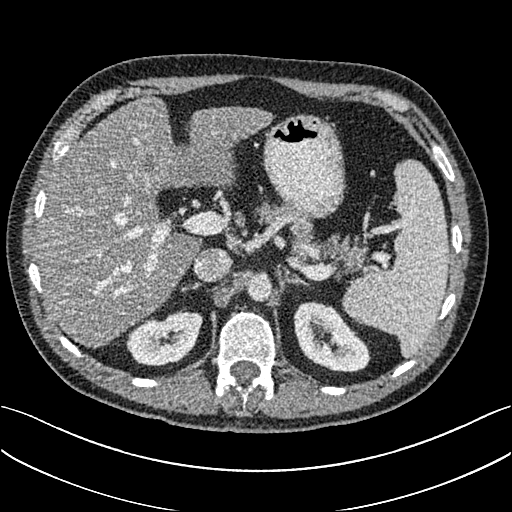}\label{fig: case21}}
\subfloat[CASE2]{\includegraphics[width=0.85in]{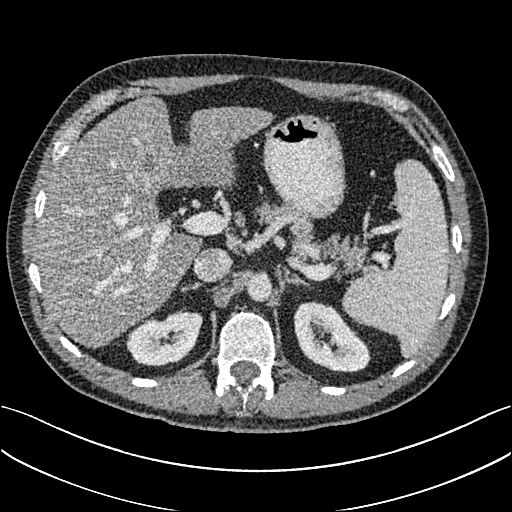}\label{fig: case22}}
\subfloat[CASE3]{\includegraphics[width=0.85in]{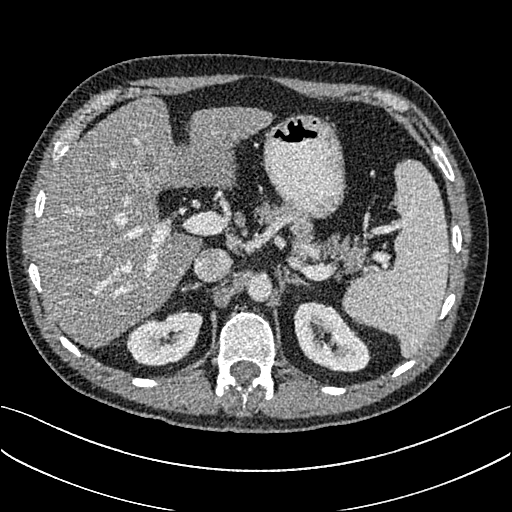}\label{fig: case23}}
\subfloat[CASE4]{\includegraphics[width=0.85in]{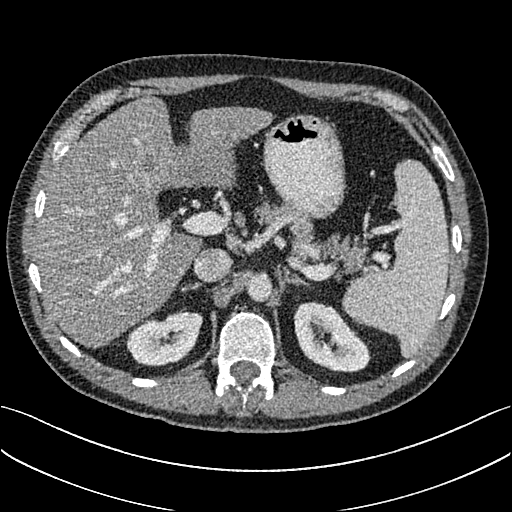}\label{fig: case24}}

\subfloat[CASE1]{\includegraphics[width=0.85in]{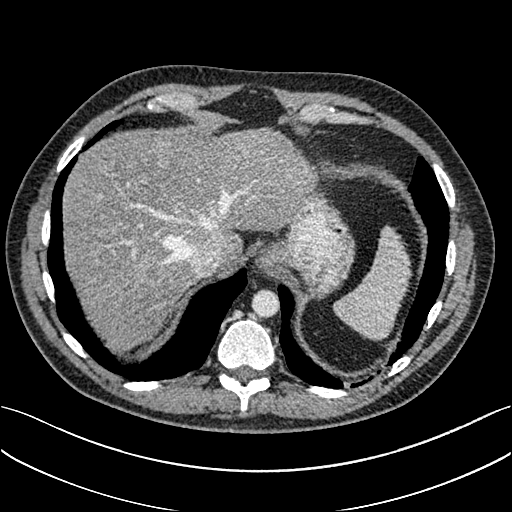}\label{fig: case31}}
\subfloat[CASE2]{\includegraphics[width=0.85in]{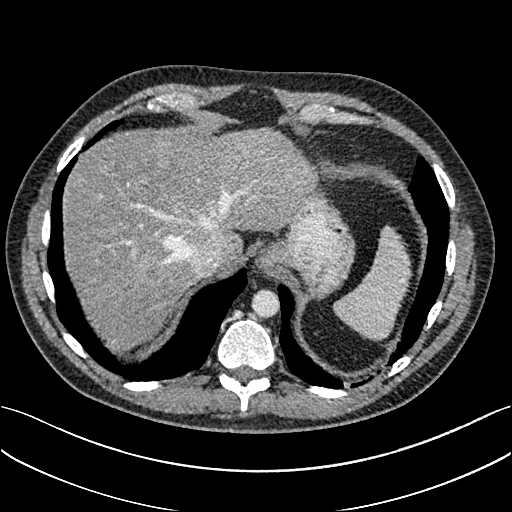}\label{fig: case32}}
\subfloat[CASE3]{\includegraphics[width=0.85in]{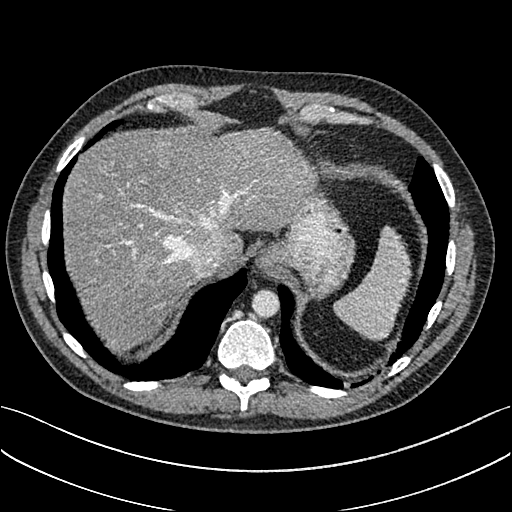}\label{fig: case33}}
\subfloat[CASE4]{\includegraphics[width=0.85in]{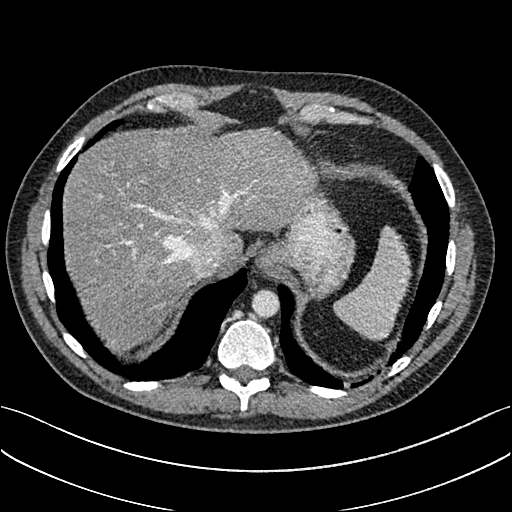}\label{fig: case34}}
\caption{Results from four different training sets for SMGAN-3D. (a)-(d) refer to Fig.~\ref{fig: example1}, (e)-(h) refer to Fig.~\ref{fig: example3} and (i)-(l) refer to Fig.~\ref{fig: example4}. This display window is [-160, 240]HU.}
\label{fig: trainingset}
\end{figure}

We randomly splitted the Mayo dataset~\cite{lowdosectgrandchallenge} into four different training sets,each with 5,000 image patches of size $80\times 80 \times 11$ pixels. Then, different training sets were used to validate the generalizability of our proposed 3D SMGAN model. The results are presented in Fig.~\ref{fig: trainingset} and Table~\ref{table: casepsnr&ssim&rmse}.
\begin{table}[!ht]
\renewcommand{\arraystretch}{1.2}
\centering
\caption{Quantitative results associated with different training sets for SMGAN-3D in Figs.~\ref{fig: trainingset}.}
\setlength{\tabcolsep}{2pt}
\begin{tabular}{c c c c c c c c c c c c}
\hline
& \multicolumn{3}{c}{Figs.~\ref{fig: case11} - \ref{fig: case14}} && \multicolumn{3}{c}{Figs.~\ref{fig: case21} - \ref{fig: case24}} && \multicolumn{3}{c}{Figs.~\ref{fig: case31} - \ref{fig: case34}}\\
& PSNR & SSIM & RMSE && PSNR & SSIM & RMSE && PSNR & SSIM & RMSE \\
\cline{2-4}\cline{6-8}\cline{10-12}
Case1 & 26.678 & 0.811 & 0.0463 && 25.842 & 0.776 & 0.0510 && 26.538 & 0.812 & 0.0472 \\
Case2 & 26.759 & 0.814 & 0.0459 && 25.848 & 0.781 & 0.0510 && 26.544 & 0.814 & 0.0470 \\
Case3 & 26.589 & 0.807 & 0.0468 && 25.701 & 0.772 & 0.0519 && 26.455 & 0.806 & 0.0475 \\
Case4 & 26.903 & 0.815 & 0.0452 && 25.914 & 0.782 & 0.0506 && 26.662 & 0.816 & 0.0464 \\
\hline
\end{tabular}
\label{table: casepsnr&ssim&rmse}
\end{table}

\section{Summary of notations}
\begin{table}[!ht]
\renewcommand{\arraystretch}{1.3}
\centering
\caption{Summary of notations.}
\begin{tabular}{p{3cm}  p{5cm} }
\hline
Notation & Meaning \\
\hline
NDCT & Normal dose CT \\
LDCT & Low dose CT \\
SSL & Structurally sensitive loss, integrating the structural loss and the $L_{1}$ loss as defined in Eq.~\ref{eq: loss_mix} \\

SSIM & Structural similarity index (SSIM)~\cite{wang2004image}  \\
MS-SSIM & Multi-scale structural similarity index (MS-SSIM)~\cite{wang2003multiscale} \\
SL-net (CNN-SL) & 8-layer CNN with only structural similarity loss\\
MSL-net(CNN-MSL) & 8-layer CNN with only multi-scale structural similarity loss\\
WGAN & Wasserstein Generative Adversarial Networks with $L_{2}$ loss\\
BM3D & Block-matching and 3D filtering\\
RED-CNN & Residual encoder-decoder CNN with only $L_{2}$ loss\\
WGAN-VGG & Wasserstein generative adversarial network with perceptual loss\\
SMGAN-2D & 2D Wasserstein generative adversarial network with SSL loss\\
SMGAN-3D & 3D Wasserstein generative adversarial network with SSL loss\\

\hline
\end{tabular}
\label{table:notations}
\end{table}

\section*{Acknowledgment}
The authors would like to thank NVIDIA Corporation for the donation of Titan Xp GPU, which has been utilized for this study. The authors are grateful for helpful discussion with Dr. Mats Persson (Stanford University). This work was supported in part by the National Natural Science Foundation of China under Grant 61671312 and Science and Technology Project of Sichuan Province of China under Grant 2018HH0070, and in part by the National Institutes of Health under Grants R21 EB019074, R01 EB016977, and U01 EB017140.

\bibliographystyle{IEEEtran}
\bibliography{denoisingCTnew}
\ifCLASSOPTIONcaptionsoff
  \newpage
\fi

\end{document}